\documentclass[10pt,twocolumn,letterpaper]{article}

\usepackage{wacv}
\usepackage{times}
\usepackage{epsfig}
\usepackage{graphicx}
\usepackage{amsmath}
\usepackage{amssymb}

\def\etal{\emph{et al. }}

\usepackage{booktabs}
\usepackage{subfigure}

\def\wacvPaperID{868} 

\wacvfinalcopy 

\ifwacvfinal
\fi

\ifwacvfinal
\usepackage[breaklinks=true,colorlinks,linkcolor=black,citecolor==black,bookmarks=false]{hyperref}
\else
\usepackage[pagebackref=true,breaklinks=true,colorlinks,linkcolor=black,bookmarks=false]{hyperref}
\fi

\begin{document}

\title{Towards Enhancing Fine-grained Details for Image Matting}

\author{Chang Liu\hspace{0.5in}Henghui Ding\footnotemark[1]\hspace{0.5in}Xudong Jiang\\
Nanyang Technological University, Singapore\\
{\tt\small \{liuc0058, ding0093, exdjiang\}@ntu.edu.sg}
}
\maketitle
\renewcommand{\thefootnote}{\fnsymbol{footnote}}
\footnotetext[1]{Corresponding author.}
\thispagestyle{empty}


\begin{abstract}
    In recent years, deep natural image matting has been rapidly evolved by extracting high-level contextual features into the model. However, most current methods still have difficulties with handling tiny details, like hairs or furs. In this paper, we argue that recovering these microscopic details relies on low-level but high-definition texture features. However, {these features are downsampled in a very early stage in current encoder-decoder-based models, resulting in the loss of microscopic details}. To address this issue, we design a deep image matting model {to enhance fine-grained details. Our model consists of} two parallel paths: a conventional encoder-decoder Semantic Path and an independent downsampling-free Textural Compensate Path (TCP). The TCP is proposed to extract fine-grained details such as lines and edges in the original image size, which greatly enhances the fineness of prediction. Meanwhile, to leverage the benefits of high-level context, we propose a feature fusion unit(FFU) to fuse multi-scale features from the semantic path and inject them into the TCP. In addition, we have observed that poorly annotated trimaps severely affect the performance of the model. Thus we further propose a novel term in loss function and a trimap generation method to improve our model's robustness to the trimaps. The experiments show that our method outperforms previous start-of-the-art methods on the Composition-1k dataset.
\end{abstract}

\section{Introduction}
\label{sec:intro}

Image matting is one of the most important tasks in computer vision community and is gaining increasing popularity in recent years. It has been widely applied in many areas, including film production, promotion image composition, and etc. The goal of image matting is to estimate the transparency, or alpha matte, of the {target} foreground object at each pixel. Mathematically speaking, a digital image $I_p$ can be formulated as a linear combination of the foreground and the background by the equation:

\begin{equation}
    I_p=\alpha_p F_p+\left(1-\alpha_p\right)B_p,\qquad\alpha_p\in \left[0,1\right]
    \label{eq:matting_def}
\end{equation}
where $F_p$ and $B_p$ denote the foreground and the background color at pixel $p$ respectively, and $\alpha_p$ is the desired alpha matte. Image matting problem is ill-posed because it targets to solve seven values ($\alpha_p$, $F_p$ and $B_p$) with only three known values ($I_p$), as seen in Eq.~(\ref{eq:matting_def}). For most of the existing approaches~\cite{xu2017deep, lu2019indices, hou2019context}, a trimap that indicates the ``pure'' foreground, ``pure'' background, and ``unknown'' region is provided with the image. It is used to reduce the size of the solution space and to indicate that which object is the target-of-interest if there are more than one foreground objects in the image.

There are many challenges in image matting. Firstly, in natural images, due to the light environment and transparency of the foreground, the color distribution of foreground and background can be very similar. This causes many traditional color-based methods suffering from severe inductive bias. Secondly, some images contain microscopic and detailed structures such as hairs or furs, which raises a challenge to the fine-grained performance of algorithms. It is easy to see that the first challenge is more related to high-level contextual features, and the second one is more related to low-level textural features that contains more spatial details. In recent years, deep learning based methods have shown their potential in image matting area. These methods not only make use of the color information of the original image but also utilize the context information inside the images, which is beneficial to the first challenge. However, for the second challenge, which requires the model to have the ability to detect very fine details, there is still much room for development.

Currently, most existing deep image matting works, e.g., ~\cite{xu2017deep, hou2019context, cai2019disentangled}, adopt the encoder-decoder architecture. However, in such architectures, the input image is downsampled in the very early stage of the network, resulting in loss of spatial details. {To restore the spatial details}, some approaches first use the encoder-decoder architecture to generate a coarse output, and then refines this coarse output using postprocessing or refinement modules such as extra convolution layers~\cite{xu2017deep} and LSTM units~\cite{cai2019disentangled} in a cascading manner. However, it is very hard to reconstruct the already lost spatial details. In this work, we propose to learn high-definition features to enhance fine-grained details for image matting. Moreover, since the aim of image matting is to regress the transparency value for each pixel, like its definition, this task is still widely considered as a low-level or mid-level computer vision task. Thus, it relies more on lower-level features than most computer vision tasks. Based on this observation, we propose that low-level but high-resolution structural features contains essential details (e.g., corners, edges, lines, etc.) and should be utilized for inferring matte estimation.

From this point of view, in this paper we propose a new deep image matting method that learns high-definition structural features and high-level contextual features in parallel. The network consists of two paths, namely Semantic Path (SP) and Textural Compensate Path (TCP).
First of all, like many previous deep image matting methods, we utilizes an encoder-decoder backbone to infer rich high-level contextual information for rough matting, {which is the semantic path in this work. Besides the semantic path, we argue that low-level but high-resolution features are desired for inferring fine-grained matte estimation. To this end, we introduce an independent downsampling-free module as textural compensate path to learn the low-level but high-definition features, which greatly enhances the fine-grained details of prediction}.

Moreover, among many test cases, especially cases with mesh-like structure such as laces and embroidery, we have observed that due to inaccurately labeled trimap, most of deep image matting algorithms cannot detect some ``absolute'' background well. To address this issue, we design a novel term in loss function and a novel trimap generation method, which enhances the performance of the network in detecting ``absolute'' backgrounds without causing overuse of video memory resource.

The major contribution of our paper can be summarized as follows:

\begin{enumerate}
    \item We present a novel {perspective} of image matting problem that explicitly divides this task into two parts: a semantic part to extract high-level semantic clues and a textural compensate part to provide fine-grained details and low-level texture clues;
    \item Based on this point we propose a new deep image matting method that explicitly defines two paths: an encoder-decoder semantic path and downsampling-free textural compensate path;
    \item We further propose a novel loss term that helps the network alleviate the inaccurately trimap issue and better detect those ``pure'' background parts;
    \item The proposed approach achieves new state-of-the-art performance on the challenging Adobe Composition-1k testing dataset.
\end{enumerate}

\section{Related Works}

Early image matting methods can mainly be divided into two categories: sampling-based approaches and propagation-based methods. Sampling-based methods, such as \cite{chuang2001bayesian,he2011global,he2013iterative,wang2005iterative,wang2007optimized}, sample and model the colors distribution inside known foreground and the background region, and the alpha value of the unknown pixel is estimated using the sampled model by some defined metrics. Recently, Tang \etal proposed a learning-based sampling approach, which introduced a neural network prior to opacity estimation and achieved remarkable performance \cite{tang2019learning}. In propagation methods, starts from the given foreground or background region, alpha values are propagated in the unknown region to background or foreground. One popular propagation-based matting method is presented in \cite{levin2007closed}.

Most current works in image matting are fully deep learning based algorithms. In order to adopt more lavish features instead of considering matting as a color problem that relies solely on color information, Xu \etal first proposed Adobe Deep Image Matting dataset, a large-scale synthetic dataset for deep learning-based matting along with a 2-step end-to-end image matting neural network \cite{xu2017deep}. The network achieved state-of-the-art performance on both synthetic images and natural images at that time. Huang \etal further designed an encoder-decoder architecture that adopts a more complex structure like residual units \cite{tang2019very}. Hou \etal designed a context-aware network composed of two encoder-decoder structures, which can recover the alpha map and foreground image at the same time \cite{hou2019context}. Cai \etal presented a deep disentangled approach that splits the image matting task into two parts: trimap adaption and alpha estimation, which is a classification task and regression task, respectively \cite{cai2019disentangled}. Zhang \etal proposed a digital matting framework in a late fusion manner that can predict the alpha matte without the input trimap \cite{zhang2019late}. Lu \etal argue that the parameter indexes in the unpooling layers in the upsampling decoder can influence the matting performance and introduced a learnable plug-in structure, namely IndexNet, to improve the pooling and upsampling process \cite{lu2019indices}. Tang \etal combined deep neural network and traditional sampling-based approaches together by using CNNs before opacity estimation \cite{tang2019learning}. Chen \etal also use a deep neural network to solve a similar problem, namely environment matting \cite{Zongker1999EnvironmentMA}.

Different from previous approaches, in this work, we argue that the low-level but high-resolution features are desired for inferring fine-grained matte estimation. We present a novel perspective of image matting problem that explicitly divides this task into two parts: a semantic part to extract high-level contextual clues and a textural part to provide fine-grained details and low-level spatial clues. The new proposed textural part greatly enhances the fineness of alpha matte prediction.

\section{Methodology}

\begin{figure*}
    \vspace{0.07in}
    \begin{center}
    \includegraphics[width=0.95\textwidth]{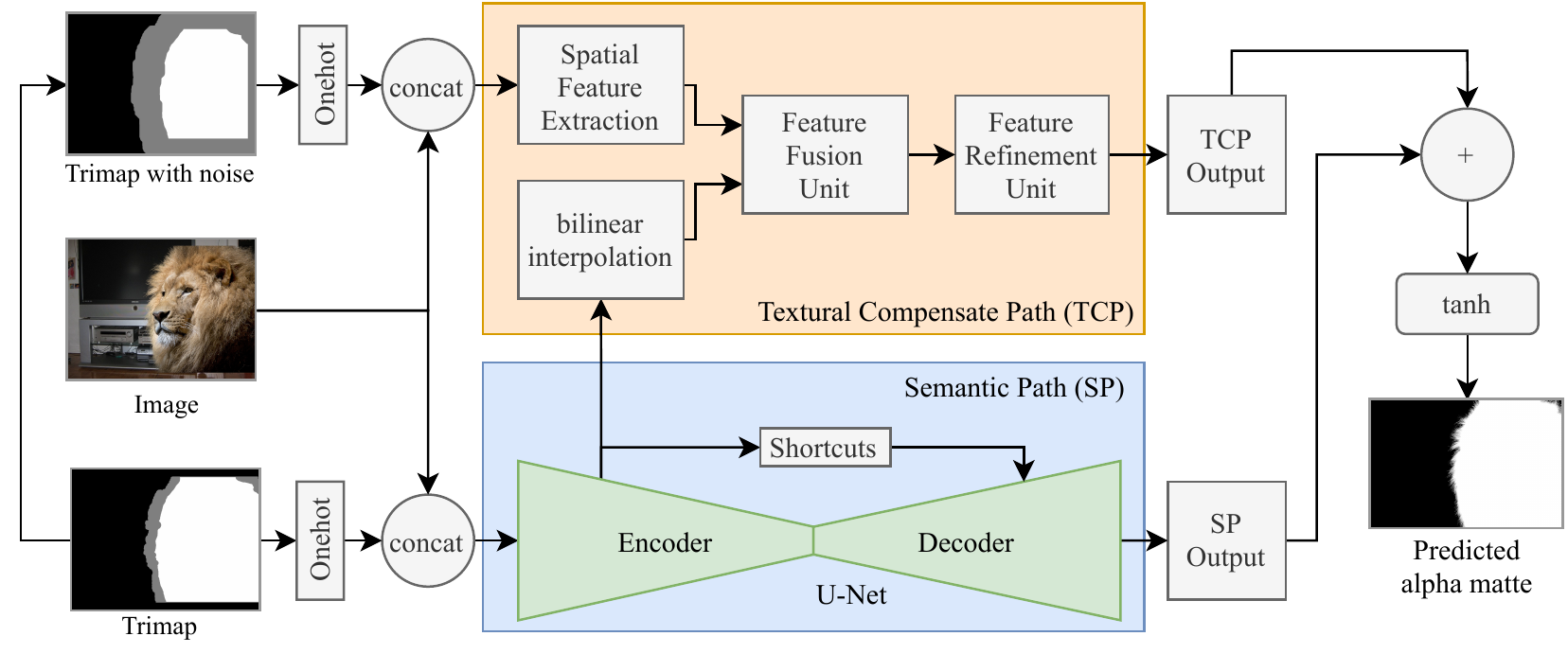}
    \end{center}
    \vspace{0.07in}
    \caption{An overview of the proposed network architecture. Our framework consists of two pathes: Semantic Path (SP) and Textural Compensate Path (TCP). The outputs of two parts are added together to form the final predicted alpha matte.}\label{fig:net_struct}
    \vspace{0.07in}
\end{figure*}

Our network consists of two parts, namely Textural Compensate Path and Semantic Path. As shown in Fig.~\ref{fig:net_struct}, the proposed network takes a 6-channel maps as the input, formed by the concatenation of the 3-channel RGB image and the corresponding one-hot encoded 3-channel trimap. The input is sent to the semantic path and textural compensate path simultaneously, where each path generates a one-channel output. Then, the $tanh$ value of the sum of the two outputs is the output of the network, i.e., the predicted alpha matte.

This section will introduce the purpose and structure of each part of the network.

\subsection{Semantic Path}

The semantic path is used to extract high-level contextual representations. The encoder-decoder architecture has been employed in many deep image matting works~\cite{xu2017deep, hou2019context,lu2019indices,cai2019disentangled} and other computer vision tasks like semantic segmentation~\cite{long2015fully, wang2019bi, ding2020scene, ding2020phraseclick, ding2020semantic, shuai2018toward} and image generation~\cite{isola2017image}. Although the size of the training dataset for image matting has been significantly increased~\cite{xu2017deep} recently, most popular datasets are synthetic, and the total number of data available is still limited compared to other computer vision tasks. Therefore, our semantic path chooses U-Net~\cite{ronneberger2015u}, which is an encoder-decoder architecture optimized for the small size of training data.

The input of the semantic path is directly taken from the network input. We slightly modify the U-Net architecture by placing two convolution layers in each shortcut to provide an adaption from low-level features to high-level features. Concretely speaking, the encoder part is built on the basis of the ResNet-34~\cite{he2016deep}, and the decoder part is built as a mirror structure of the encoder. The semantic path itself is able to work independently, and can also produce remarkable results, so that we also use the stand-alone semantic path as our baseline model.

\subsection{Textural Compensate Path (TCP)}

\begin{figure}
    \begin{center}
    \includegraphics[width=0.45\textwidth]{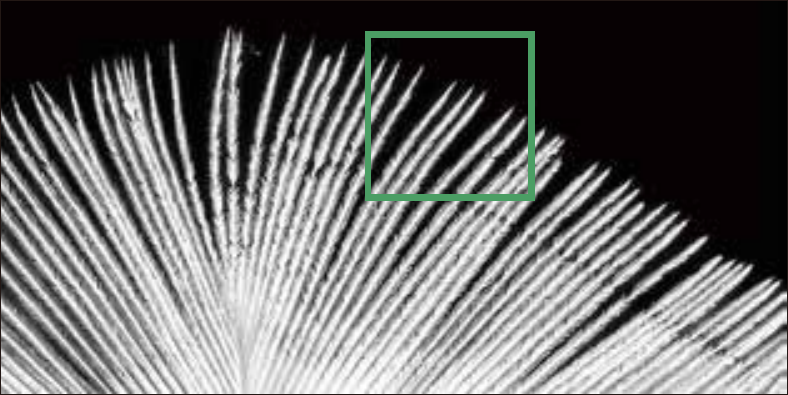}

    \vspace{0.07in}
    \includegraphics[width=0.225\textwidth]{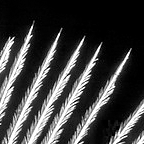}
    \includegraphics[width=0.225\textwidth]{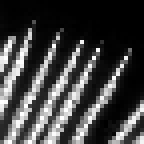}
    \end{center}
    \vspace{0.07in}

    \caption{Qualitative example for downsampling an image and reconstruct it again using upsampling methods. Fine-grained spatial details are severely lost in such process even the downsampling rate is small ($4\times$). Accurate alpha matting will be almost impossible to generate using the reconstructed image. Top: image overview; Bottom: a cropped patch of the original image and $4\times $ downsampled then $4\times$ upsampled image.}
    \label{fig:resize}
\end{figure}

As discussed in Section \ref{sec:intro}, low-level but high-resolution features that carry textural details are vital to image matting task, but these features are severe damaged due to the early downsampling in many existing encoder-decoder based approaches. To qualitatively demonstrate this issue, we randomly select a testing image from the Adobe Deep Image Matting Dataset~\cite{xu2017deep}, as shown in Fig.~\ref{fig:resize}. In the figure, we first downsample the images by 4 times and then recover the image to the original size using nearest interpolation, which is the same with our baseline model do, to demonstrate the detail loss in encoder-decoder models. It can be seen that a considerable amount of details and fine-grained structures are lost after the reconstruction. Although some previous encoder-decoder based works have explored to restore the lost details by postprocessing or refinement modules such as extra convolution layers~\cite{xu2017deep} and LSTM units~\cite{cai2019disentangled} in a cascading manner, it is very hard to reconstruct the already lost fine-grained details. In addition, some simply stacked strutures will bring extra difficulties on the training of the network, e.g. making the network cannot be trained end-to-end.

\begin{figure*}
    \vspace{0.05in}
    \begin{center}
    \includegraphics[width=0.95\textwidth]{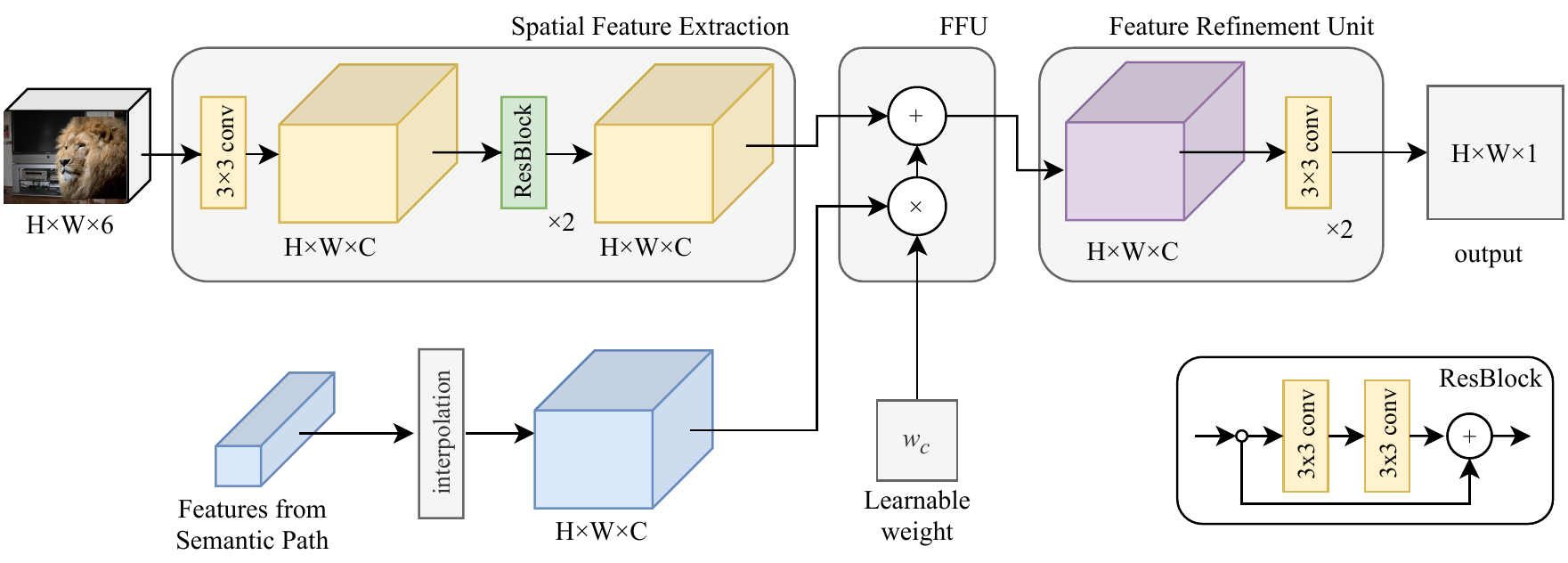}
    \end{center}
    \caption{The detailed structure of our Textural Compensate Path (TCP). The original image is first sent to Spatial Feature Extraction Unit. Then the output is sent to the Feature Fusion Unit (FFU) together with resized intermediate features from Semantic Path. Next, the Feature Refinement Unit futher extract the useful information of the output of FFU, and generate the output of the TCP. The ``$\times n$'' means that the part inside the box is duplicated for $n$ times.}\label{fig:spat_path}
    \vspace{0.1in}
\end{figure*}

To address this issue, we propose to keep as many spatial details as possible for image matting. To this end, we design a dedicated downsampling-free Textural Compensate Path (TCP) for extracting pixel-to-pixel high-definition information from features whose size is as the same as the original image, aiming to compensate for the lost pixel-to-pixel feature caused by the early downsampling in the encoder-decoder architecture in the Semantic Path. {Besides the high-resolution, another benefit is that the Textural Compensate Path learns low-level structural features, which provide low-level texture clues (e.g., edges, corners, etc) and help to estimate alpha matte in microscopic details. The architecture of this path is show in Fig.~\ref{fig:spat_path}, it consists of 3 parts}: the first part is the spatial feature extraction unit, which is formed by one convolution layer followed by two residue blocks, aiming to extract rich pixel-level structural features. This module is downsampling-free, resulting in the output size to be $H\times W$. At the same time, intermediate features from the Semantic Path are taken and resized to $H\times W$, the same as the output of the spatial feature extraction unit. Next, these two sets of features are sent to the Feature Fusion Unit (FFU). This step is to provide multi-scale and pretrain information in addition to the pixel-level spatial features. Then, fused features are sent to the feature refinement unit that consists 2 convolution layers, generating the output of TCP.

\paragraph{Feature Fusion Unit (FFU).} Though the primary purpose of the Textural Compensate Path is to extract pixel-level structural features. However, multi-scale and pretrain features are also beneficial for generating robust output. In order to introduce multi-scale features while keeping the parameter size controllable, we borrow the intermediate features from the semantic path as multi-scale features. At the same time, to ensure that the Textural Compensate Path focuses on low-level features, features are taken from very shallow layer: the second layer in the U-Net semantic path, for fusion. The features are firstly resized to the original image size using nearest interpolation. Since the feature representation in two paths can also be very different, simply adding the features from different path can be harmful to training. Thus, as shown in Fig.~\ref{fig:spat_path} we multiply the weight from semantic path by a learnable weight $w_c$ to control its influence.

\subsection{Improving Model's Robustness to Trimaps}\label{sec:improve_trimap}

\begin{figure}
    \begin{center}
    \includegraphics[width=0.155\textwidth]{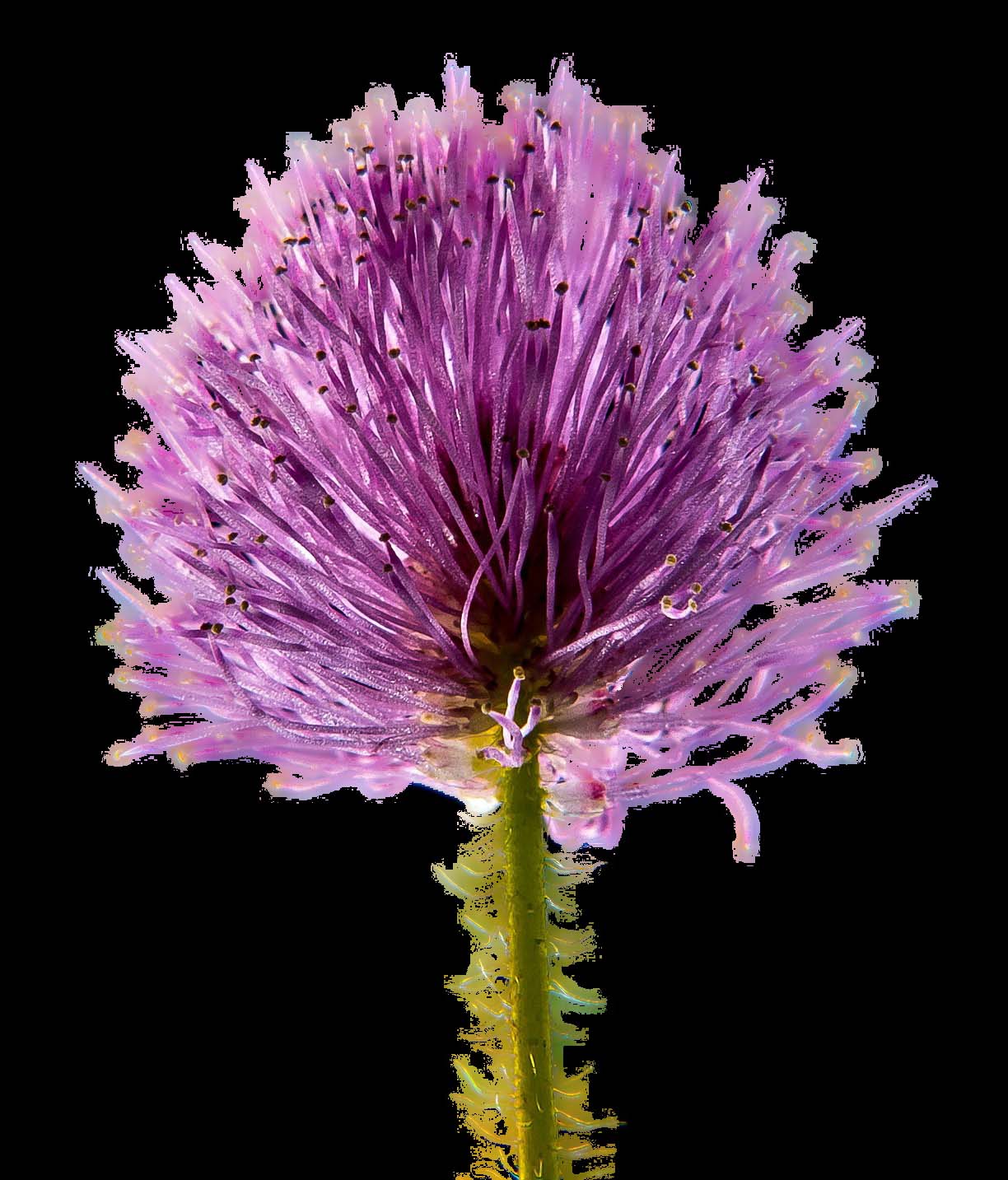}
    \includegraphics[width=0.155\textwidth]{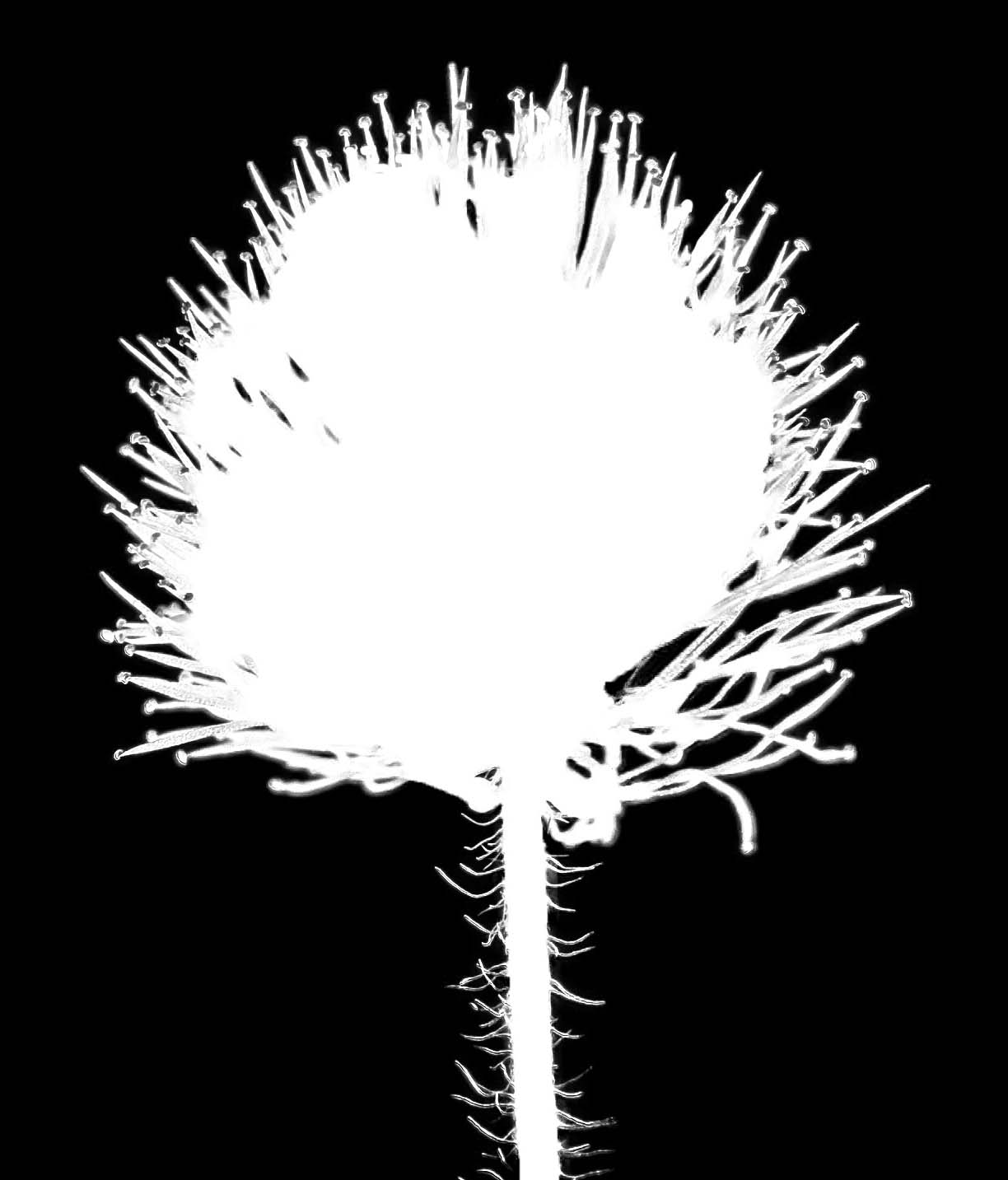}
    \includegraphics[width=0.155\textwidth]{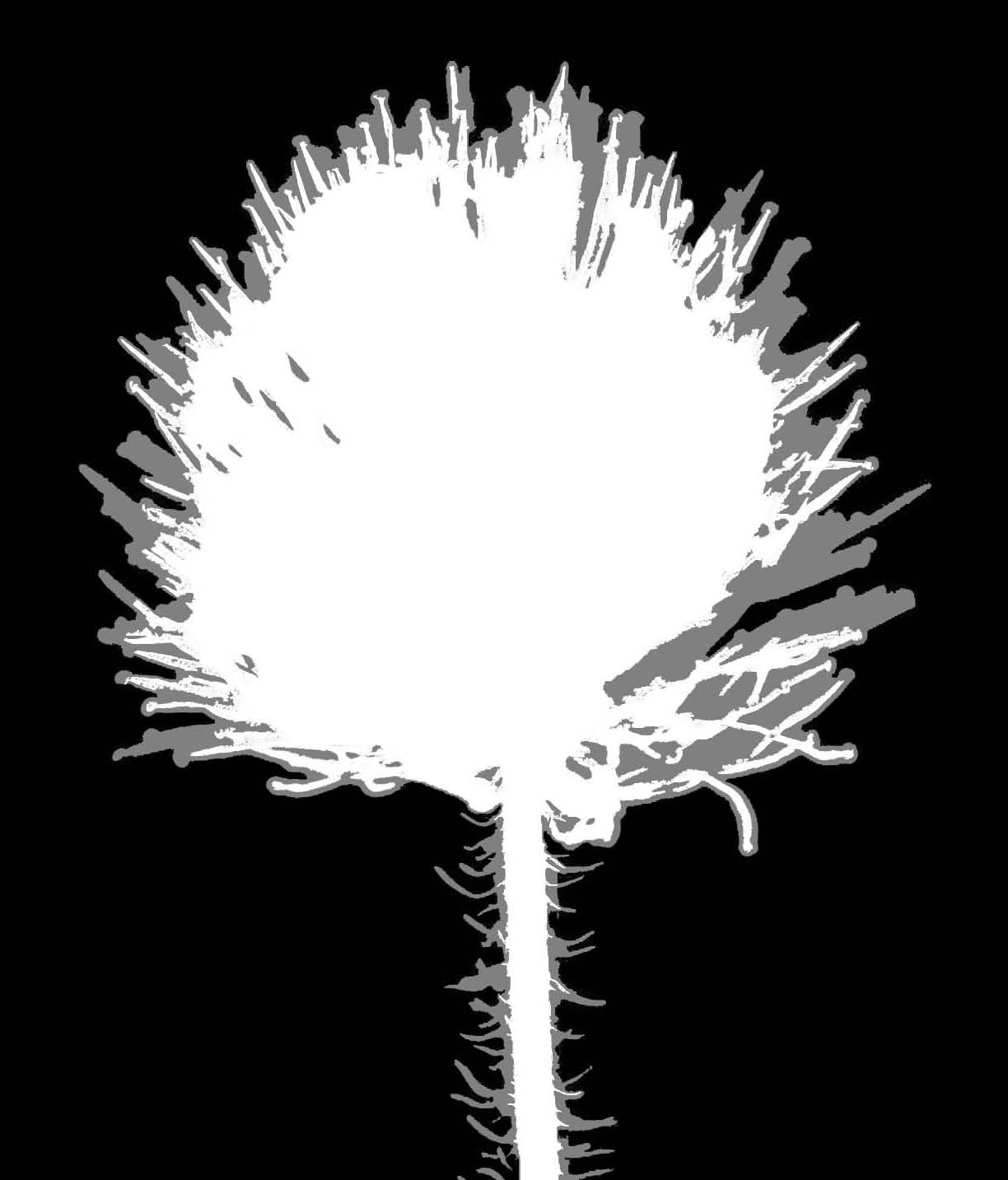}

    \vspace{0.05in}
    \includegraphics[width=0.155\textwidth]{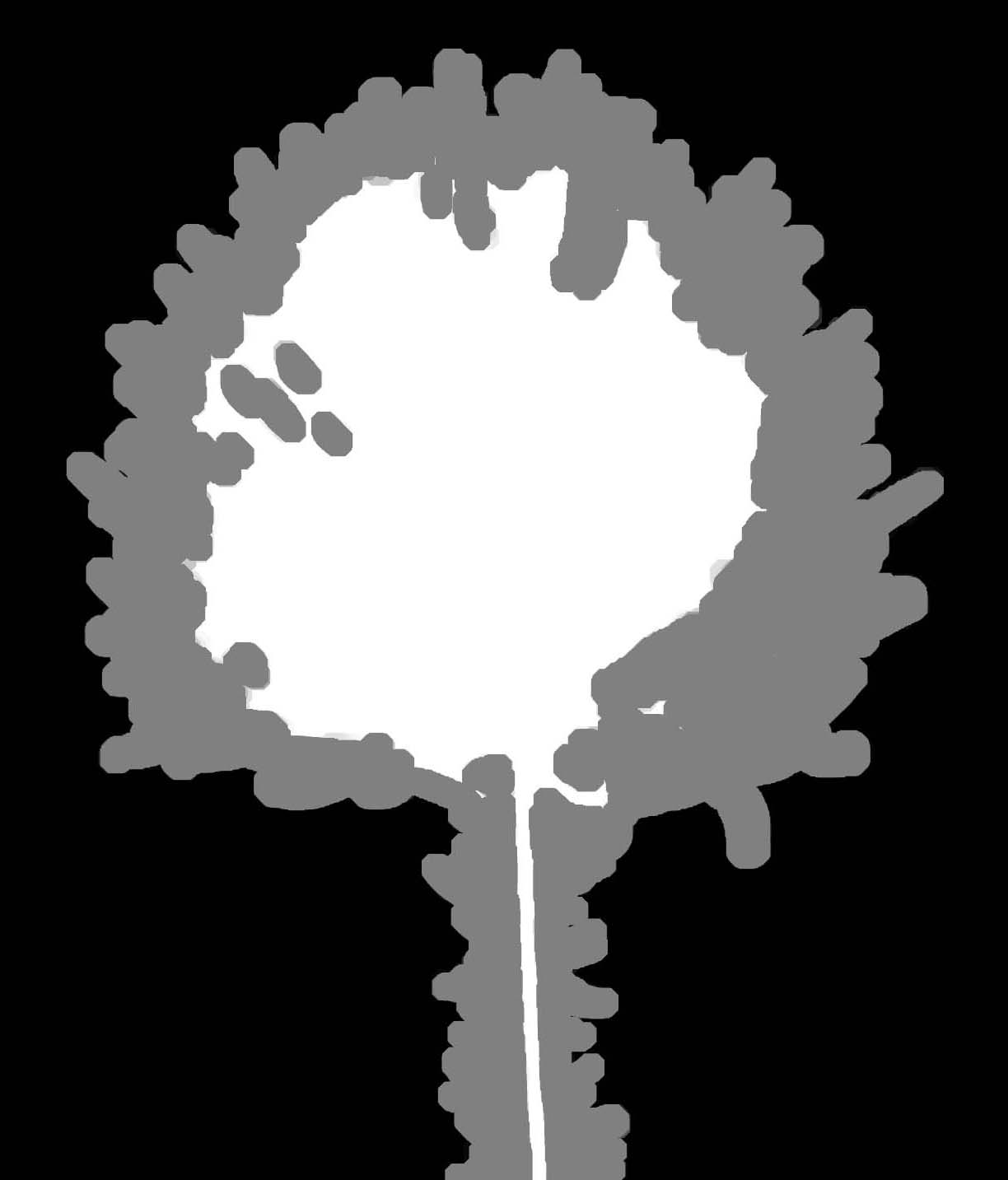}
    \includegraphics[width=0.155\textwidth]{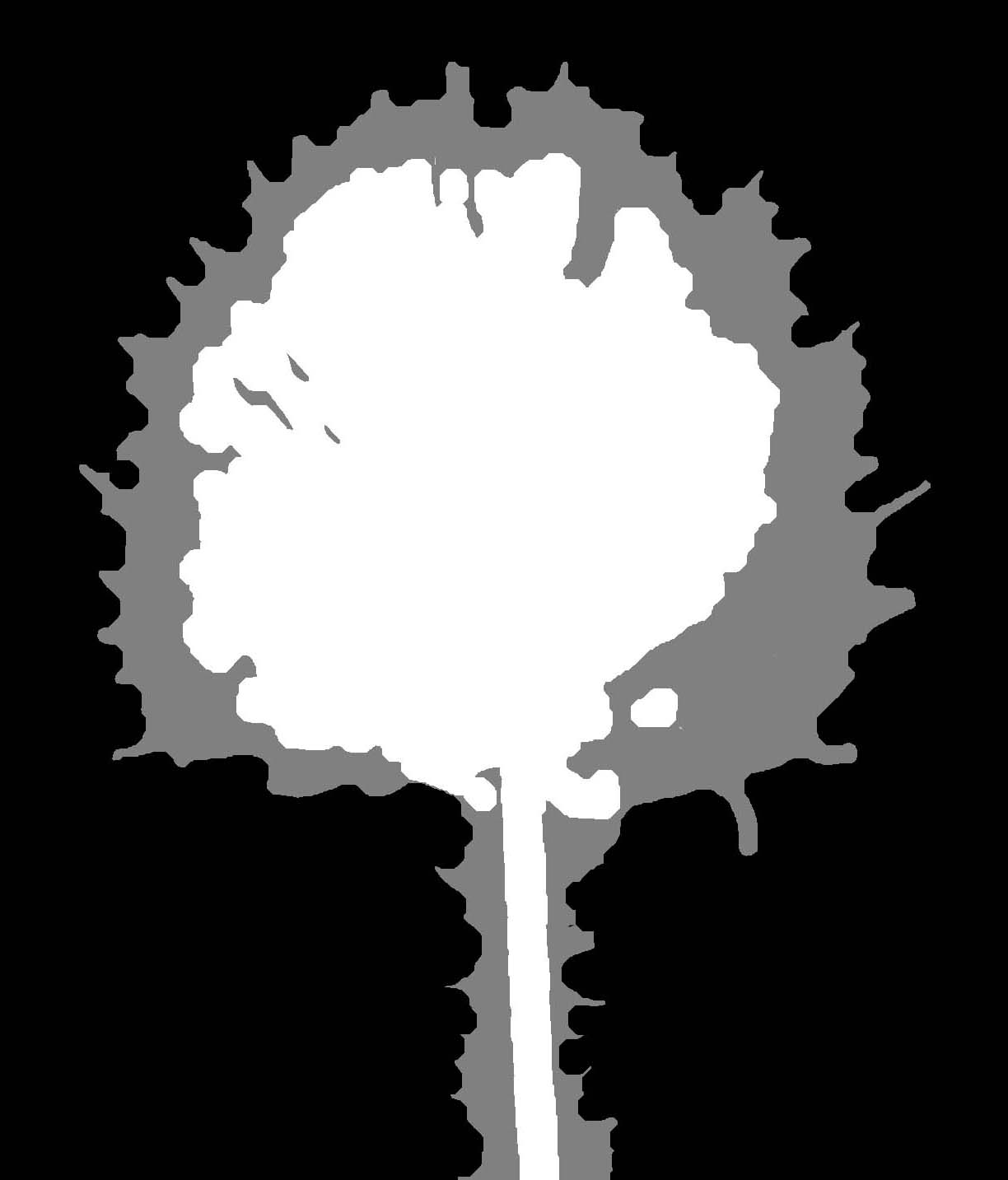}
    \includegraphics[width=0.155\textwidth]{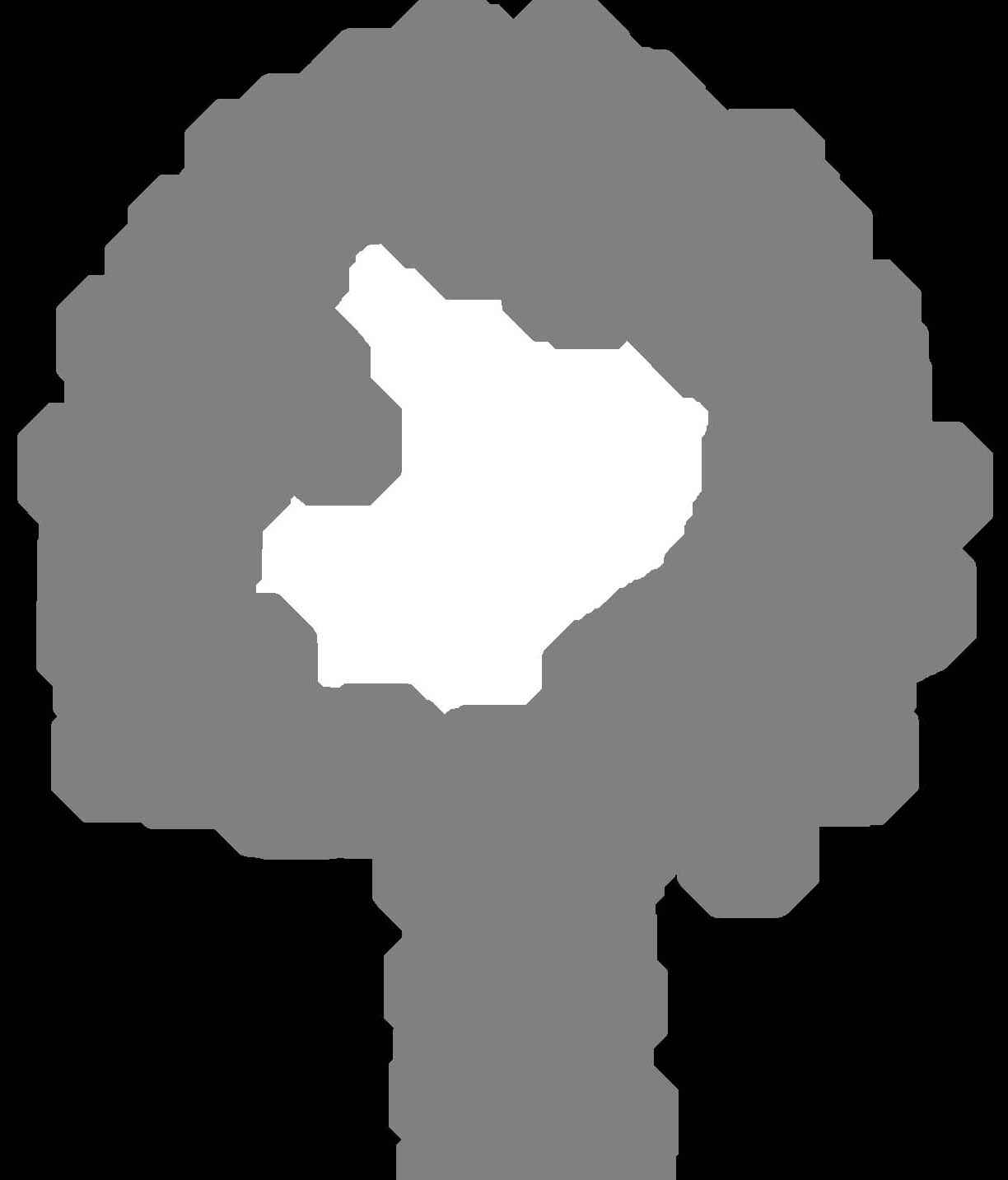}
    \end{center}
    \caption{Examples of our trimap generation method. From left to right, top: image, alpha matte, trimap generated by Eq.~\ref{eq:trimap}; bottom: generated trimap for semantic path, two trimaps generated for textural compensate path.}
   
    \label{fig:tm_gen}
\end{figure}

We have also observed another challenging part caused by the noisy trimap. As mentioned in Section \ref{sec:intro}, Trimap is usually supposed to be provided by users. However, the currently most widely-used dataset Adobe Deep Image Matting Dataset does not provide trimaps for training and requires models to generate trimap by themselves. In practice, the user-provided trimaps might be very coarse because annotating trimap is a very bothering process, especially for unprofessional users. We have observed that for a number of images in the Composition-1k testing set, nearly the whole trimaps are annotated as ``unknown region'', which means that the trimap is very coarse and almost cannot provide any useful interaction information. In contrast, for training set, model-generated trimaps are usually based on the ground-truth alpha map, resulting in very high quality. This causes the {inconsistencies} between training and testing. Here we propose two methods to give the model more robustness on handling different kinds of trimaps.

\paragraph{Trimap generation.} Currently, most approaches generate the training trimap according to the corresponding ground-truth alpha map. Concretely speaking, the trimap of point $p$ is first decided by its corresponding alpha value $\alpha_p$:
\begin{equation}
    trimap_p=\left\{
    \begin{aligned}
         & Foreground & \alpha_p=1   \\
         & Unknown    & 0<\alpha_p<1 \\
         & Background & \alpha_p=0
    \end{aligned}
    \right.
    \label{eq:trimap}
\end{equation}
Then the unknown region is enlarged by eroding foreground and background regions. However, the dilemma is that a large erosion kernel will be harmful to the network to learn context information; yet a small kernel will make the inconsistencies between training and testing trimaps larger. Thus, we take one step further. In our network, when training, trimaps are first generated by the process mentioned above with a relatively small erosion kernel size (randomly chosen between $1\times 1$ and $30\times 30$) to keep more contextual information. This trimap is used as a part of the input of the semantic path. Next, we apply extra $n$ steps of random morphological operations to the unknown region of the semantic path trimap to simulate the randomicity in noisy trimaps provided by users. Each step is randomly chosen from a $p$-iteration erosion and a $p$-iteration dilation, where $n$ and $p$ are random numbers between 0 and 3. For each step, the kernel size is randomly chosen from $1\times 1$ to $30\times 30$ for dilation and from $1\times 1$ to $10\times 10$ for erosion. This noisier trimap is used as the input of the textural compensate path. Then when inferring, the user-provided trimap is used for both paths. Some examples are shown in Fig.~\ref{fig:tm_gen}. This process endows our model more robustness when handling trimaps in different qualities.

\paragraph{Loss Function} The main loss function used in our network is the alpha-prediction loss introduced by Xu \etal~\cite{xu2017deep}. The loss is simply the absolute difference between the ground truth and predicted alpha map in each pixel. The loss value across the image is formulated as:
\begin{equation}
    L_a = \frac{1}{N_U}\sum_{p \in U}\sqrt{(\alpha_p - \hat{\alpha_p})^2 + \epsilon^2}
\end{equation}
where $U$ is the ``unknown'' region annotated in the trimap, $N_U$ is the number of pixels inside region $U$. $\alpha_p$ and $\hat{\alpha_p}$ is the ground-truth and predicted alpha value of pixel $p$. $\epsilon$ is a small positive number to guarantee the full expression differentiable.

\begin{figure}
    \begin{center}
    \subfigure[Ground-Truth alpha matte]{
        \begin{minipage}[t]{0.46\textwidth}
            \centering
            \fbox{\includegraphics[height=1.32in]{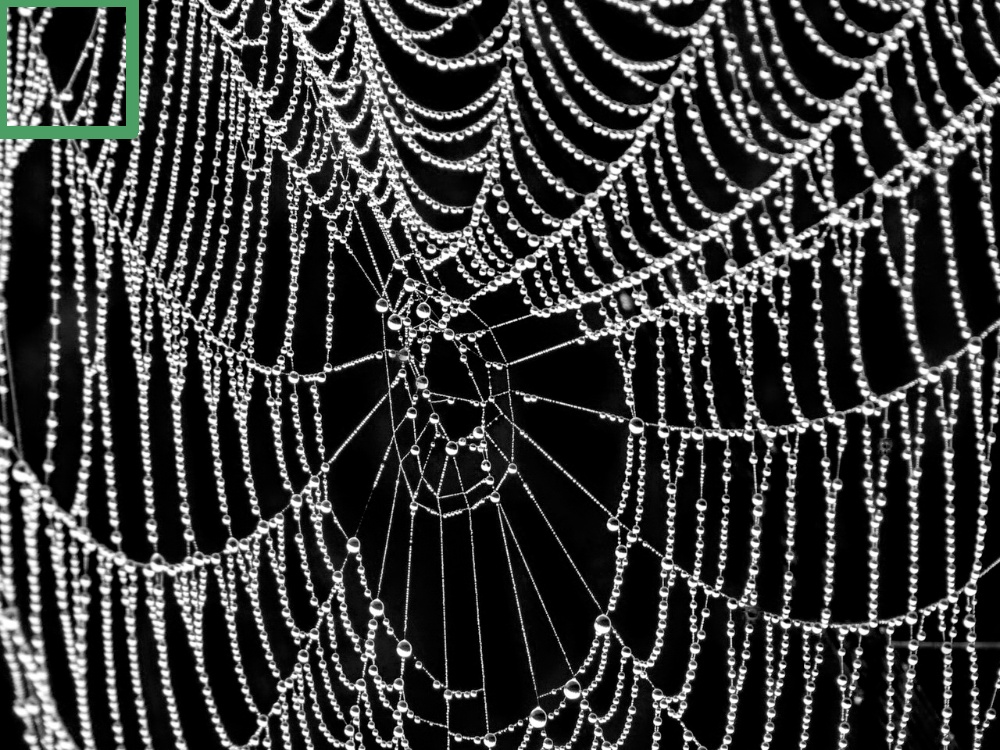}}
            \fbox{\includegraphics[height=1.32in]{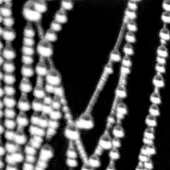}}
        \end{minipage}
    }
    \vspace{0.07in}

    \subfigure[Predicted alpha matte]{
        \begin{minipage}[t]{0.46\textwidth}
            \centering
            \fbox{\includegraphics[height=1.32in]{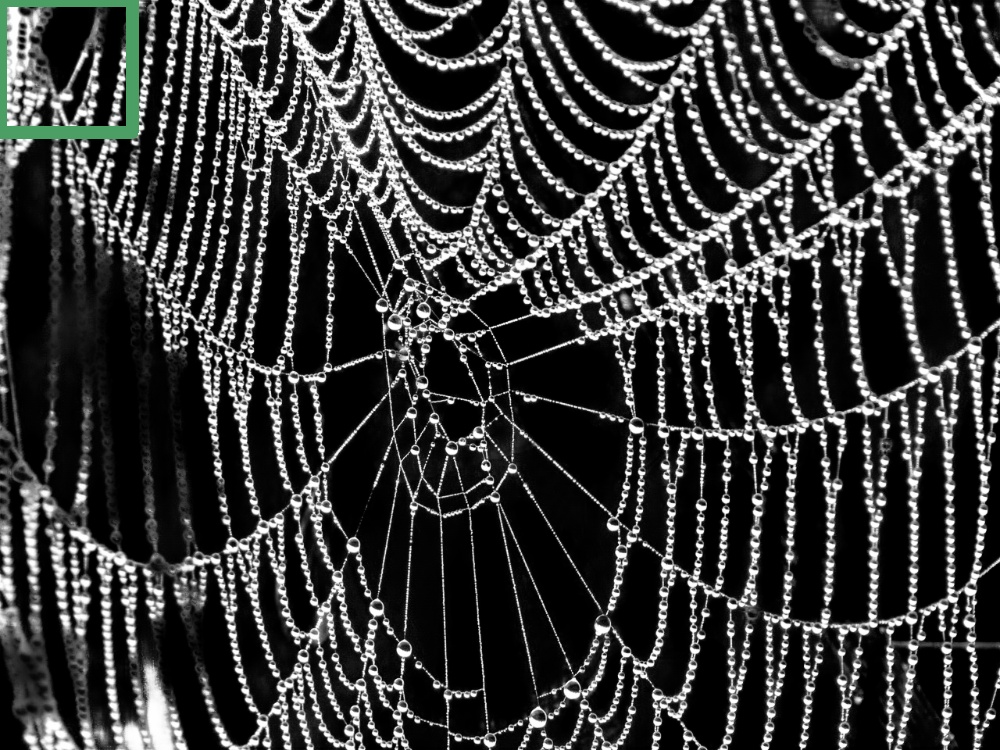}}
            \fbox{\includegraphics[height=1.32in]{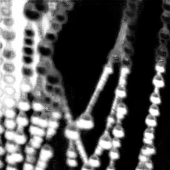}}
        \end{minipage}%
    }
    \end{center}
    \vspace{0.07in}
    \caption{Alpha matte predicted by our baseline model and ground-truth. Some areas that supposed to be ``Absolute'' backgrounds are contrastly predicted as ``Absolute'' foregrounds. The trimap of the whole region is annotated as ``unknown''.}
    \label{fig:loss}
\end{figure}

One thing to notice here is that this alpha-prediction loss only considers the unknown region in the trimap and ignores the contents in the absolute foreground and background regions. This characteristic makes the network easier to train because it reduces the solution space by filling the absolute background and foreground with value 0 or 1 according to the trimap after prediction. However, this brings a significant drawback: lots of contextual information are lost, causing the network hard to handle the ``pure'' background inside the unknown region, as shown in Fig.~\ref{fig:loss}. Some works address this issue by deriving a more accurate trimap~\cite{cai2019disentangled}. However, this will bring extra complexities to the network design. Instead, we propose another auxiliary loss, Background Enhancement Loss. This loss term recognizes the ``pure'' background inside the unknown region, and make use of these areas to give contextual guidance for the network. Our Background Enhancement Loss is defined as follows:
\begin{equation}
    \begin{aligned}
        L_{bg} & = \frac{1}{N_{bg}} \sum_{p \in R_{bg}}\sqrt{(\alpha_p - \hat{\alpha_p})^2 + \epsilon^2} \\
        R_{bg} & = \left\{\alpha_p < \theta,\quad \forall p \in U\right\}
    \end{aligned}
\end{equation}
where $R_{bg}$ is the ``absolute'' background part inside the unknown region; $N_{bg}$ is the number of pixels of $R_{bg}$, and $\theta$ is the background threshold to control the size of $R_{bg}$. The full loss of the network is then a weighted sum of the two loss terms: $L = w_1\cdot L_a+ w_2\cdot L_{bg}$. In our settings, we fix $w_1=0.9, w_2=0.1$ and $\theta=0.1$. Note that though currently the dataset we used is synthetic, only images and trimaps that are already synthesized are used in training. This makes our network available to work on both synthetic and real-world datasets.

\section{Experiments}


\subsection{Implementation Details}

The encoder of our model is first initialized with a pretrained ResNet-32 model on ImageNet \cite{imagenet_cvpr09} and then trained end-to-end on the Adobe Deep Image Matting Dataset. The network is trained for 300,000 steps in total including 7500 warmup steps~\cite{goyal2017accurate}. The Adam optimizer~\cite{kingma2014adam} with $\beta_1=0.5$ and $\beta_2=0.999$ is used for training the network. Inspired by \cite{loshchilov2016sgdr}, we apply the cosine annealing training strategy with the initial learning rate $\lambda=4\times 10^{-4}$. Our model is trained on four 8GiB Nvidia GTX1080s with the batch size of 24 in total. All convolution layers in the textural compensate path except the output layer, are followed with a ReLU activation function and a synced batch normalization. Full size images in Adobe Deep Image Matting Dataset can be inferred on one single Nvidia GTX1080 card.

We test the performance of our model on the Composition-1k dataset \cite{xu2017deep}.The Composition-1k testing set has 50 unique foreground objects. 1000 testing samples is sampled by using 20 randomly chosen backgrounds from Pascal VOC~\cite{pascal-voc-2012} for each foreground. We use the original code provided in \cite{xu2017deep} for both generating test samples and evaluating performance.

\subsection{Ablation Study}

\begin{table}
    \vspace{0.1in}
    \begin{center}
        \begin{tabular}{|l|rrrr|}
            \hline
            Method            & SAD           & MSE            & Grad          & Conn          \\
            \hline\hline
            Baseline          & 42.1          & 0.011          & 20.4          & 40.7          \\
            Baseline+TCP      & 38.8          & 0.011          & 19.2          & 36.4          \\
            Baseline+TCP+IMRP & \textbf{37.6} & \textbf{0.009} & \textbf{18.3} & \textbf{35.4} \\
            \hline
        \end{tabular}
    \end{center}
    \vspace{0.07in}

    \caption{Quantitative comparisons to the baseline on Composition-1k dataset. ``IMRP'' represents Improving Model’s Robustness to Trimaps in Section \ref{sec:improve_trimap}}
    \label{tab:ablation}
    \vspace{0.1in}
\end{table}

\begin{figure*}
    \begin{center}
    \vspace{0.1in}
    \begin{minipage}[t]{\textwidth}
        \centering
        \fbox{\includegraphics[width=0.24\textwidth]{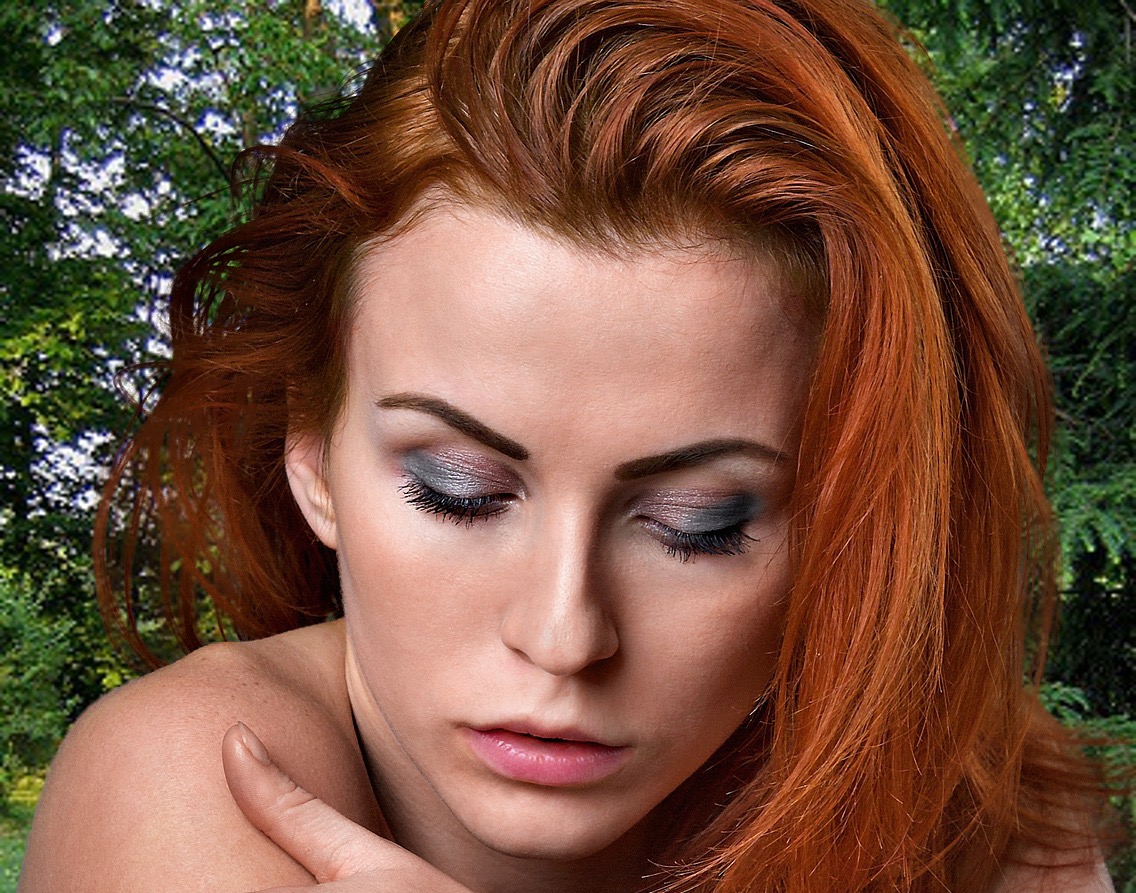}}\hspace{0.03in}
        \fbox{\includegraphics[width=0.24\textwidth]{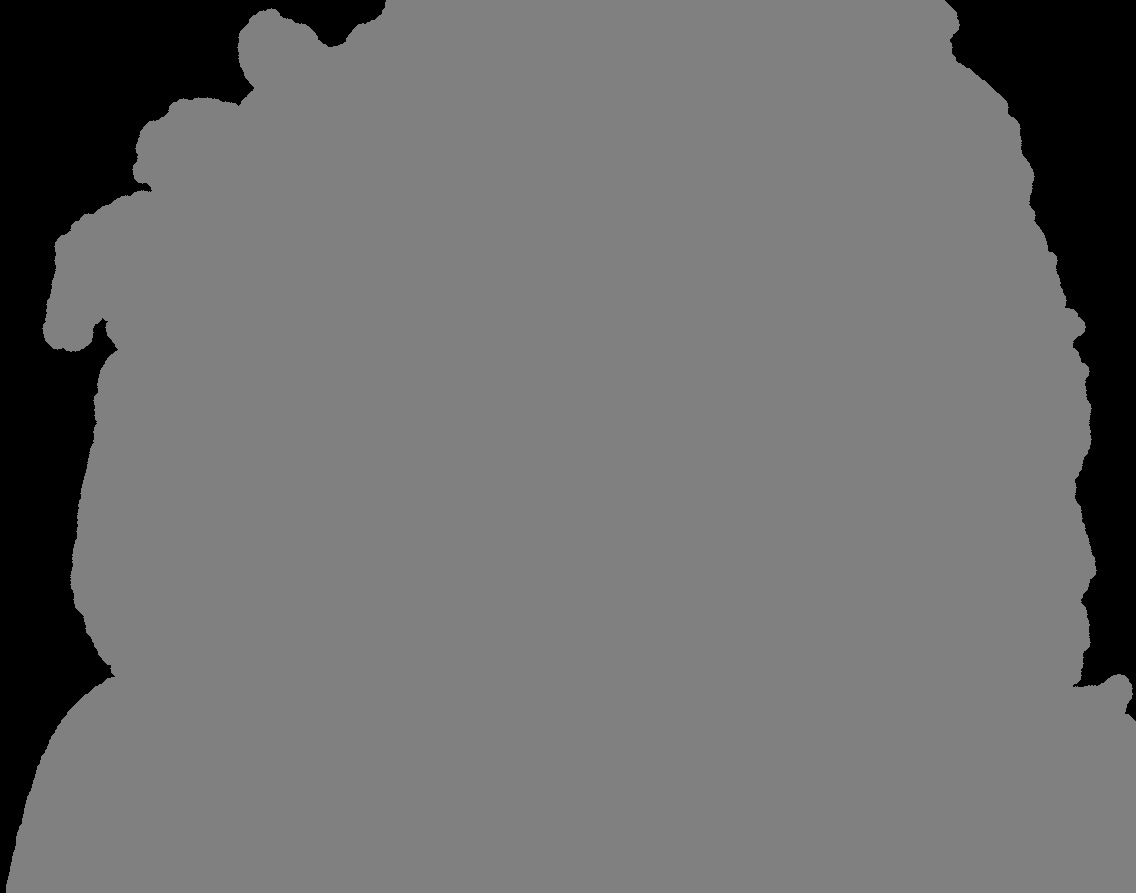}}\hspace{0.03in}
        \fbox{\includegraphics[width=0.24\textwidth]{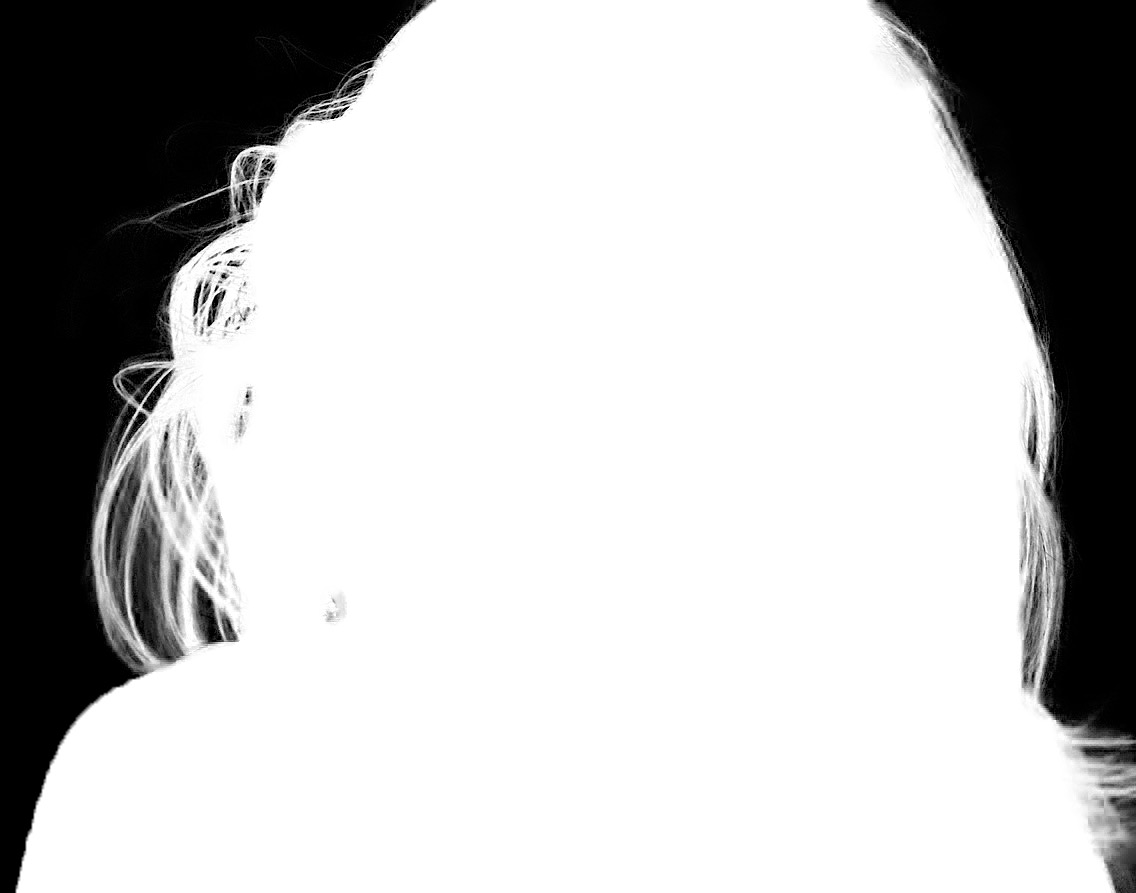}}\hspace{0.03in}

        \vspace{0.05in}
        \fbox{\includegraphics[width=0.24\textwidth]{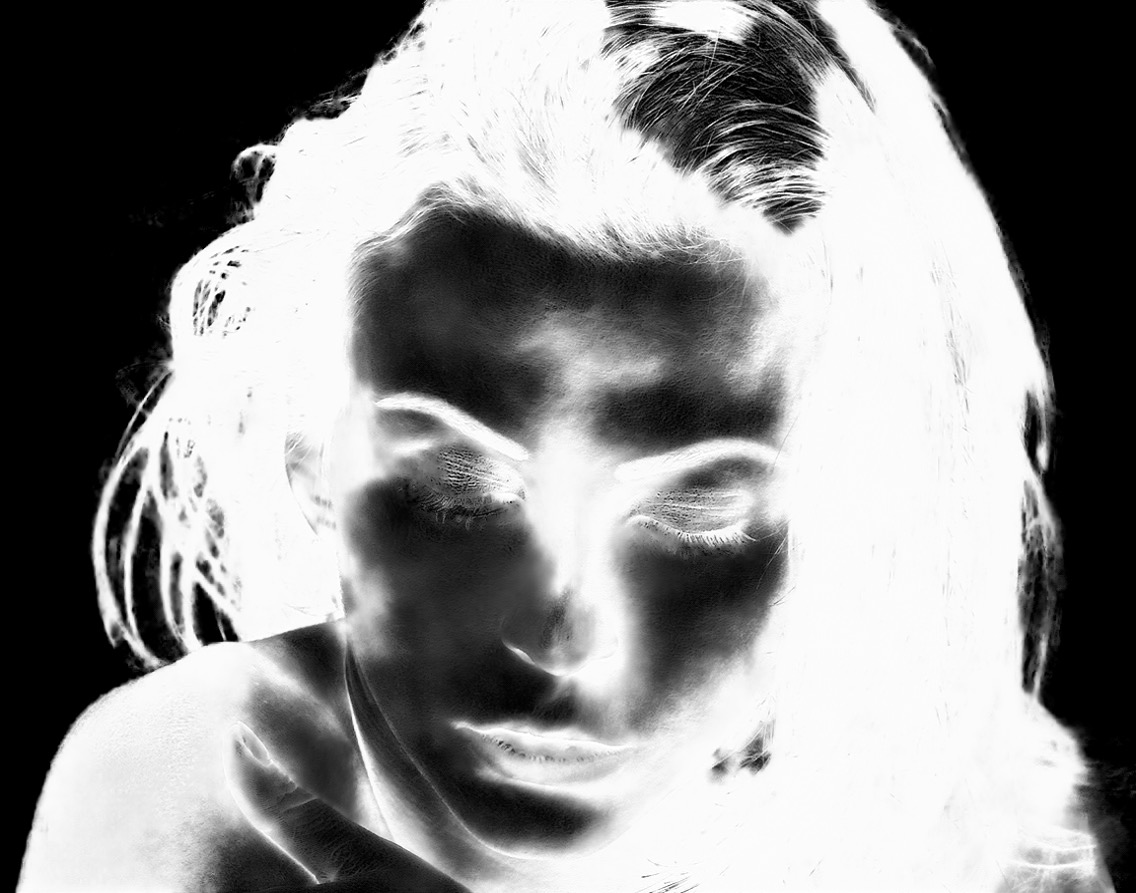}}\hspace{0.03in}
        \fbox{\includegraphics[width=0.24\textwidth]{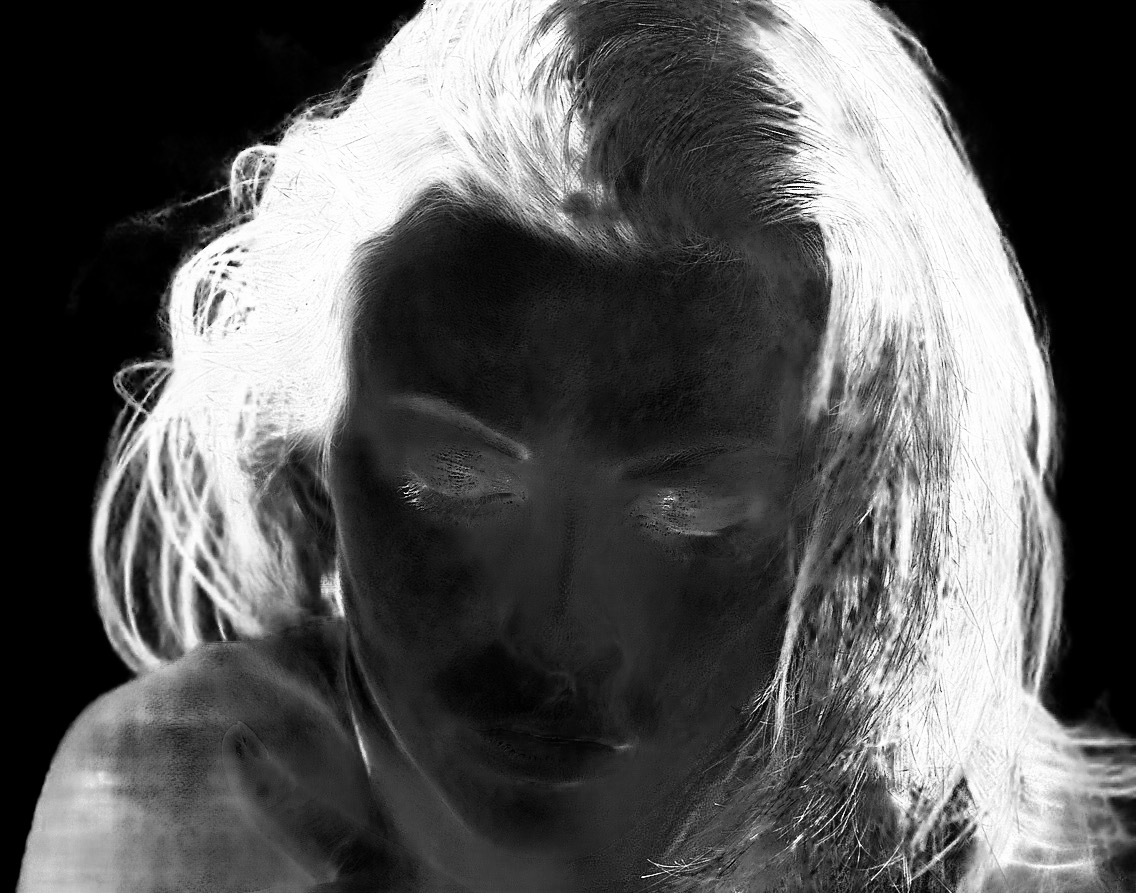}}\hspace{0.03in}
        \fbox{\includegraphics[width=0.24\textwidth]{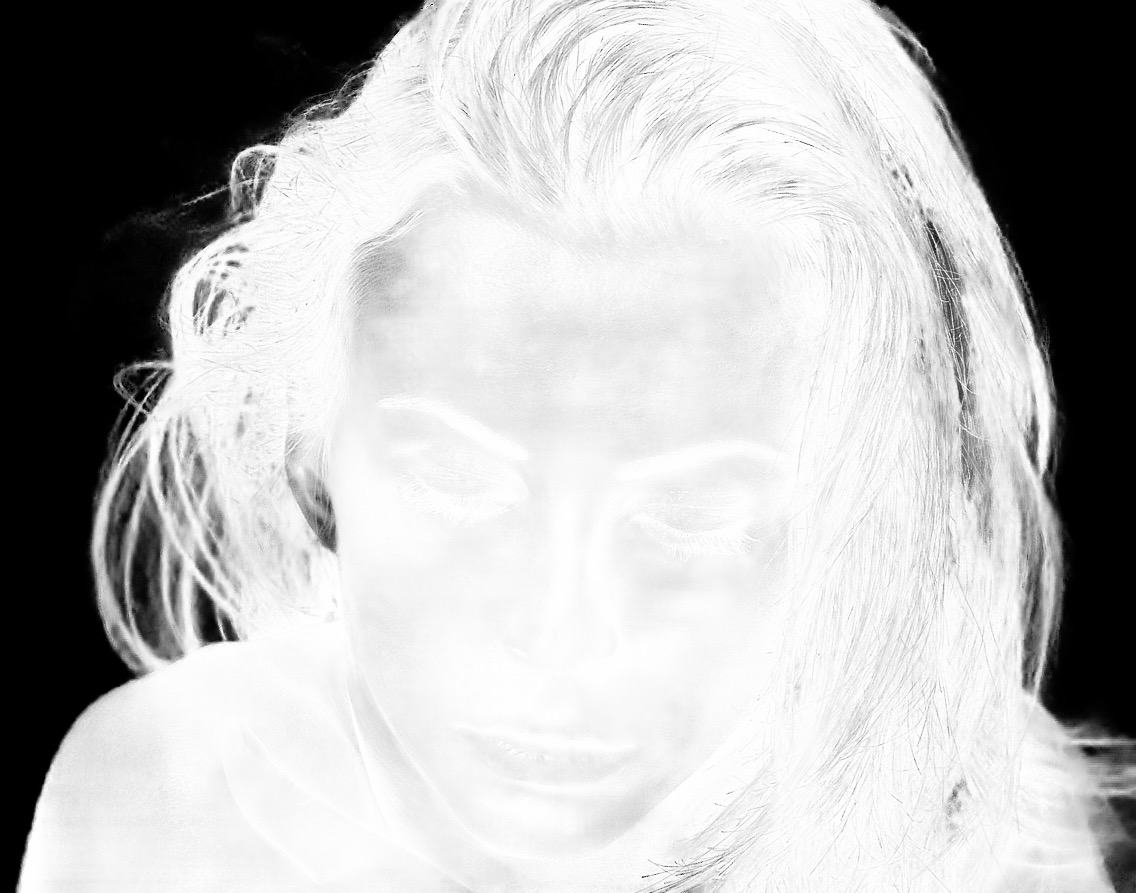}}\hspace{0.03in}
    \end{minipage}%
    \end{center}
    \vspace{0.07in}
    \caption{Example results of our method on image with low-quality trimap. The trimap contains no ``foreground'' region. From left to right: original image, trimap, ground-truth, IndexNet~\cite{lu2019indices}, our baseline, our proposed.}
    
    \label{fig:results_trimap}
\end{figure*}

\begin{table}
    \begin{center}
        \begin{tabular}{|l|rrrr|}
            \hline
            Methods                                    & SAD           & MSE            & Grad          & Conn          \\
            \hline\hline
            Global Matting \cite{he2011global}         & 133.6         & 0.068          & 97.6          & 133.3         \\
            Closed-Form \cite{levin2007closed} & 168.1         & 0.091          & 126.9         & 167.9         \\
            KNN Matting \cite{chen2013knn}             & 175.4         & 0.103          & 124.1         & 176.4         \\
            Deep Matting \cite{xu2017deep}       & 50.4          & 0.014          & 31.0          & 50.8          \\
            AdaMatting \cite{cai2019disentangled}      & 41.7          & 0.010          & \textbf{16.8} & -             \\
            IndexNet \cite{lu2019indices}              & 45.8          & 0.013          & 25.9          & 43.7          \\
            SampleNet \cite{tang2019learning}          & 40.4          & 0.010          & -             & -             \\
            \hline

            Ours                                       & \textbf{37.6} & \textbf{0.009} & 18.3          & \textbf{35.4} \\
            \hline
        \end{tabular}
    \end{center}
    \vspace{0.07in}
    \caption{Testing results on the Composition-1k Dataset. The best results in each metric is emphasized in bold. For all metrics, smaller is better.}
    \label{tab:result}
\end{table}

In this section we report the ablation study results to show the effectiveness of all parts of our network. The results are listed in Table \ref{tab:ablation}. Besides the popular SAD (Sum of Absolute Distance) and MSE (Mean Squared Error) metrics, we also use another two metrics: Gradient and Connectivity \cite{rhemann2009perceptually} proposed by Rhemann \etal to evaluate the perceptually matting performance. It can be seen from the Table \ref{tab:ablation} that the performance of the model with Textural Compensate Path is significantly improved in terms of all metrics compared to the baseline model. This proves that the TCP successfully extracted useful features that are lost in the baseline encoder-decoder model. Moreover, our trimap generation methods and the novel Background Enhancement Loss further improved the overall results. Besides, to show the effectiveness of our trimap generation method more clearly, we give an intuitionistic demonstration of a test case whose trimap is poorly annotated, as shown in Fig.~\ref{fig:results_trimap}. In this case, the whole foreground region in the given trimap are annotated with ``unknown'', which means the trimap do not provide any information on the ``absolute foregrounds''. Our model successfully detected foreground regions without the direct guidance of the trimap.

\subsection{Experiment Results}

In this section we compare the performance of our model with other state-of-the-art works. We compare our result with 3 non-deep-learning methods: Global Matting~\cite{he2011global}, Closed-Form Matting~\cite{levin2007closed} and KNN Matting~\cite{chen2013knn} as well as 4 deep learning based methods: Deep Image Matting~\cite{xu2017deep}, AdaMatting~\cite{cai2019disentangled}, IndexNet~\cite{lu2019indices} and SampleNet~\cite{tang2019learning}.

\begin{figure*}[t]
    \centering

    \subfigure[``cobweb'']{
        \begin{minipage}[t]{0.99\textwidth}
            \centering
            \fbox{\includegraphics[width=0.24\textwidth]{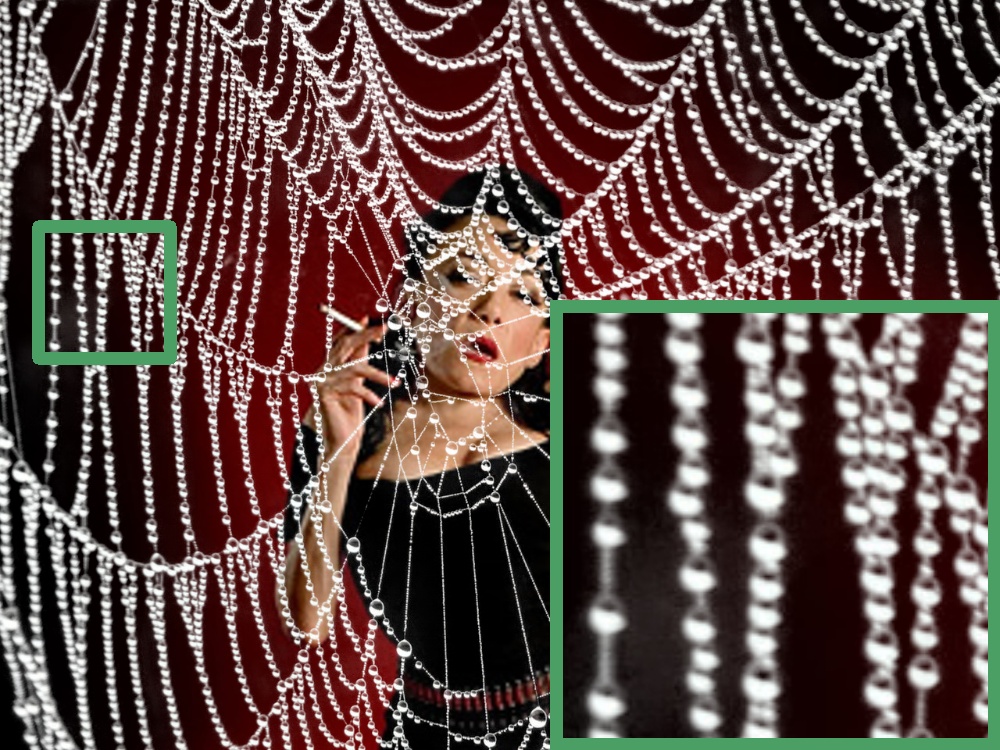}}
            \fbox{\includegraphics[width=0.24\textwidth]{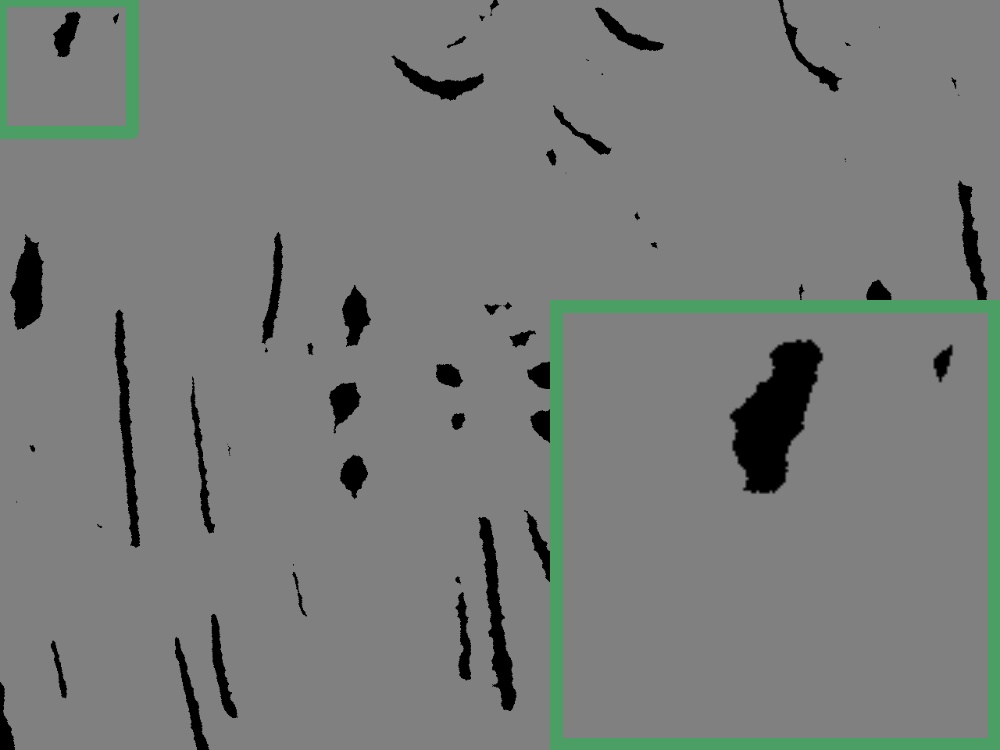}}
            \fbox{\includegraphics[width=0.24\textwidth]{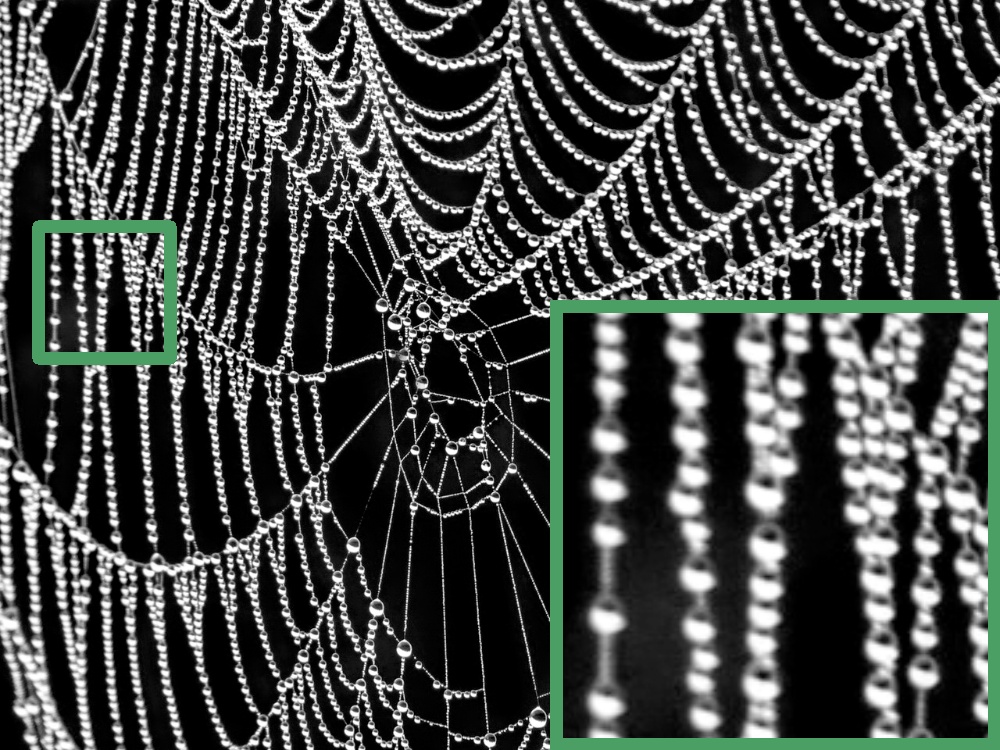}}
            \fbox{\includegraphics[width=0.24\textwidth]{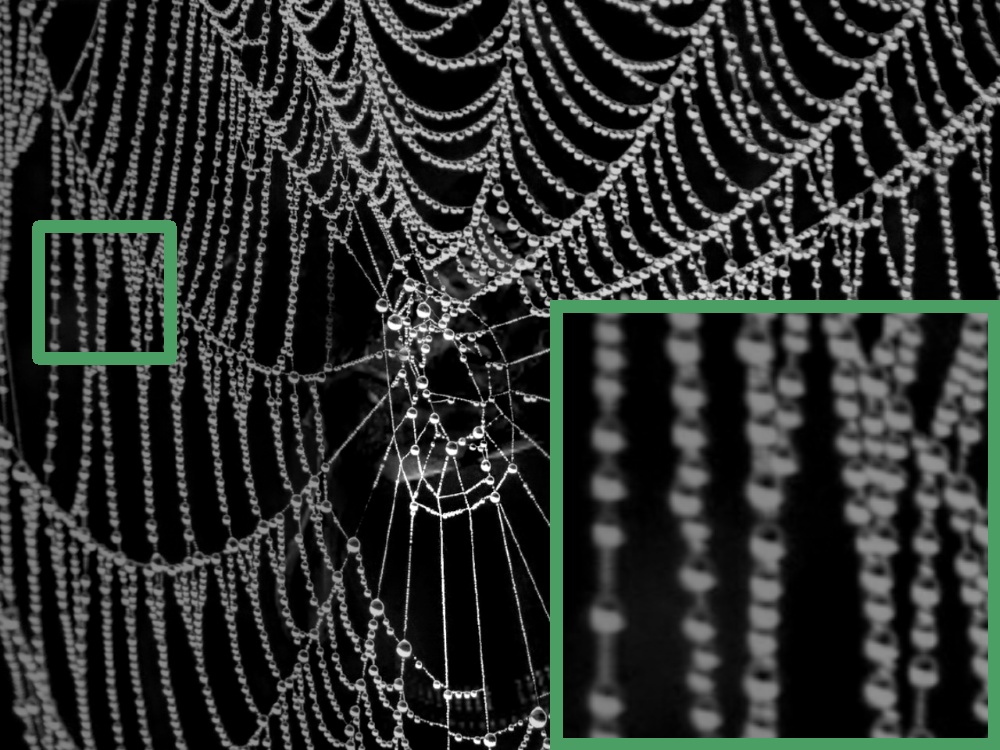}}

            \fbox{\includegraphics[width=0.24\textwidth]{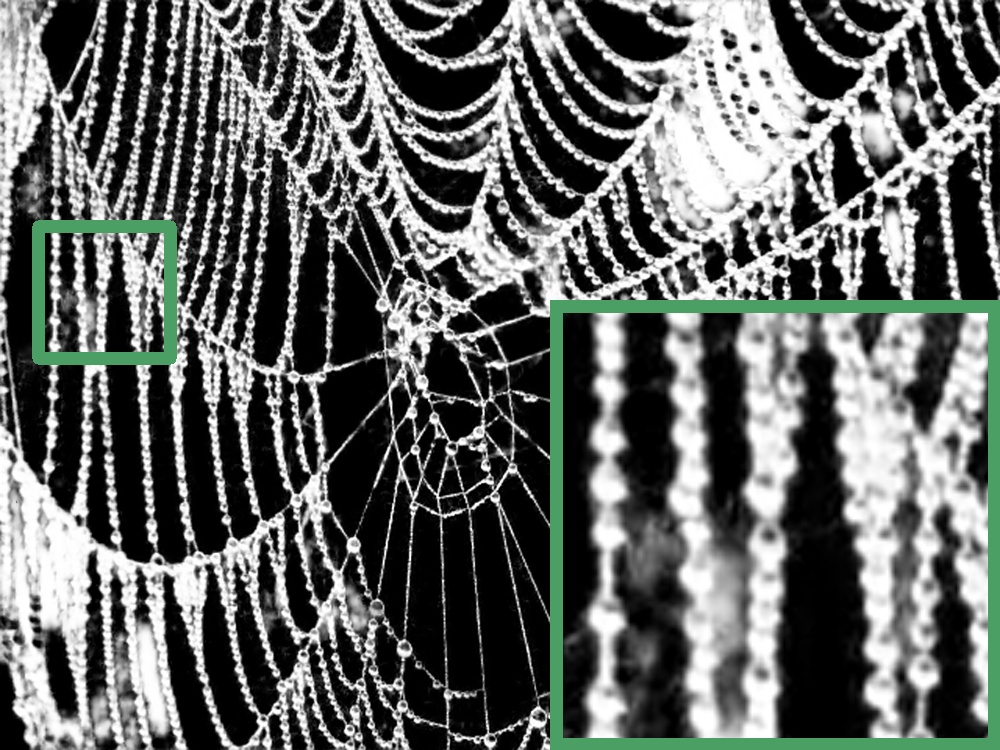}}
            \fbox{\includegraphics[width=0.24\textwidth]{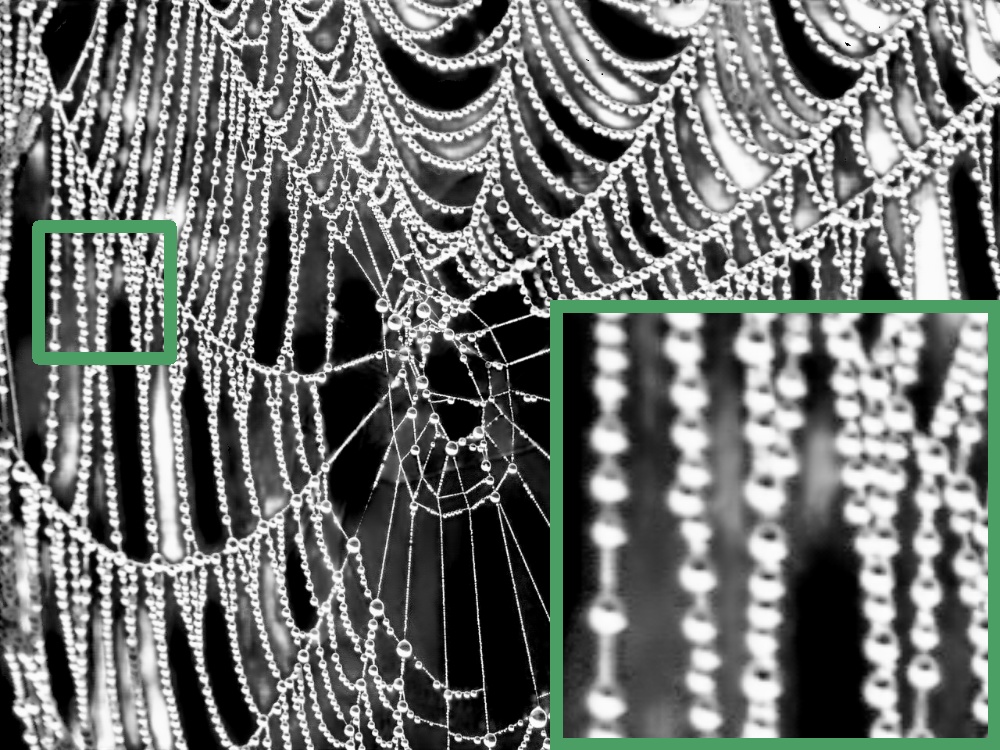}}
            \fbox{\includegraphics[width=0.24\textwidth]{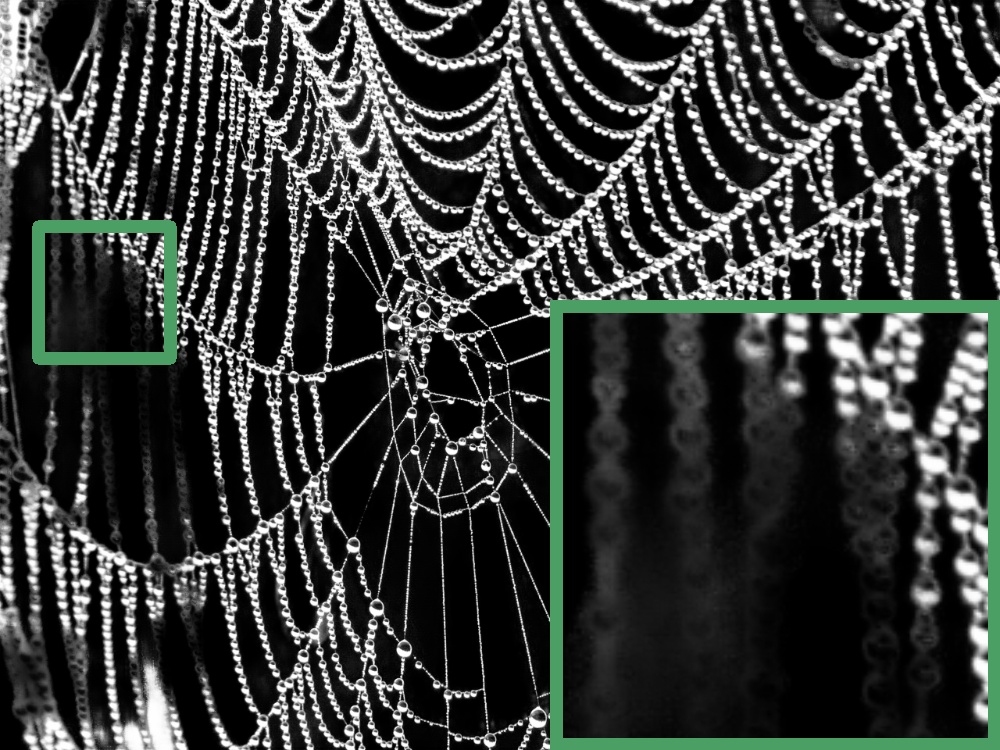}}
            \fbox{\includegraphics[width=0.24\textwidth]{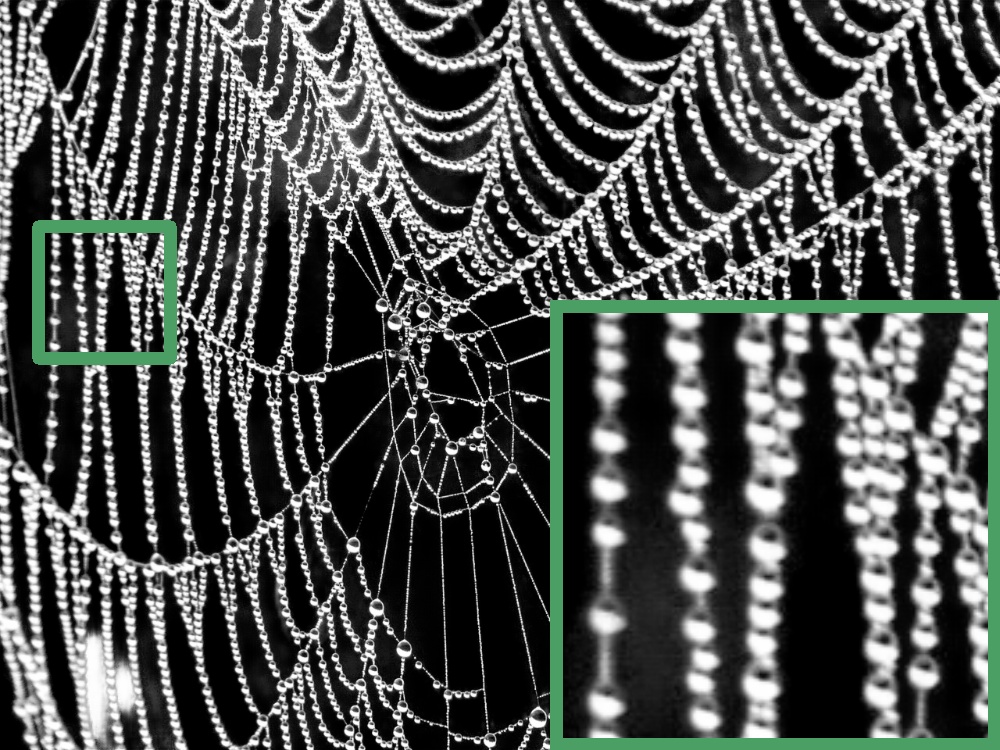}}
            \vspace{0.04in}
        \end{minipage}%
    }%

    \subfigure[``spring'']{
        \begin{minipage}[t]{0.99\textwidth}
            \centering
            \fbox{\includegraphics[width=0.24\textwidth]{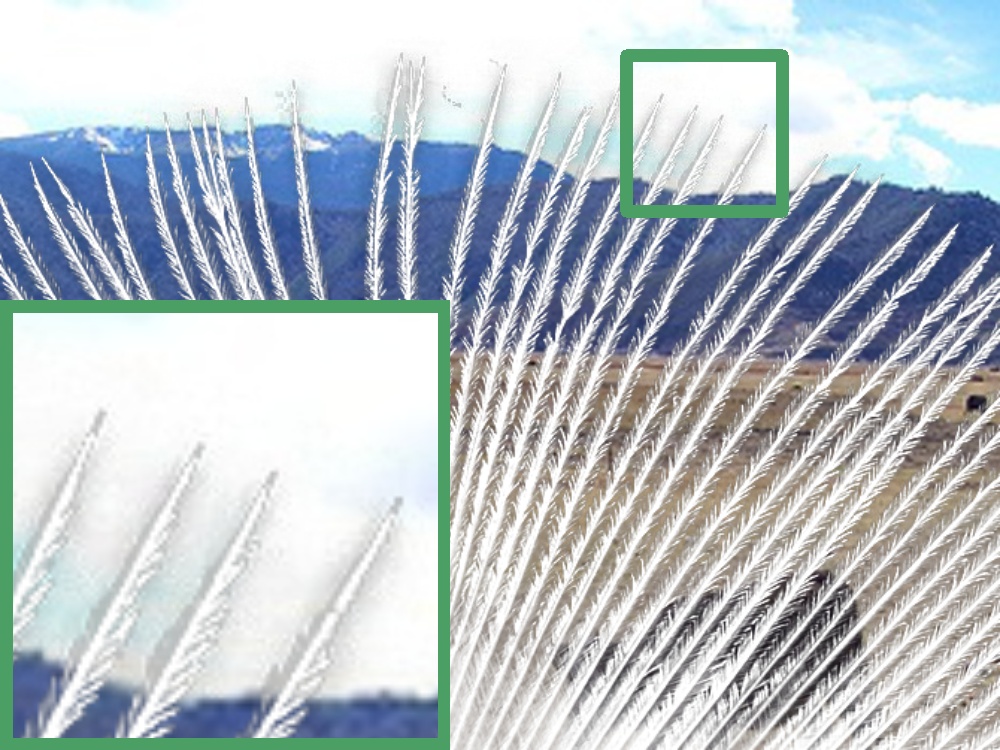}}
            \fbox{\includegraphics[width=0.24\textwidth]{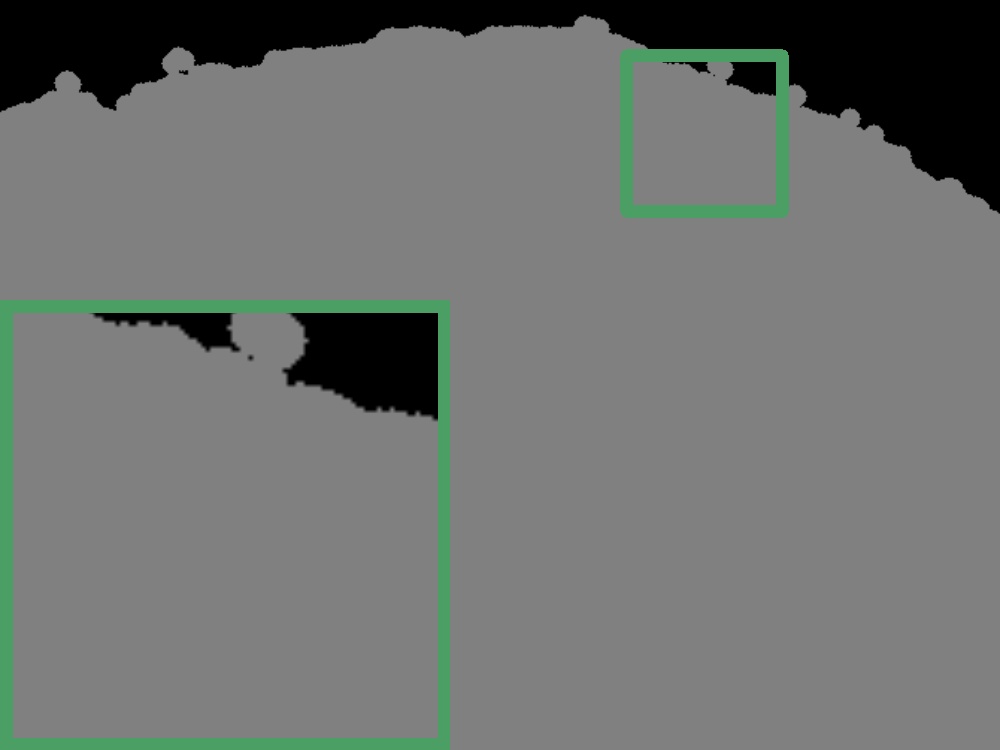}}
            \fbox{\includegraphics[width=0.24\textwidth]{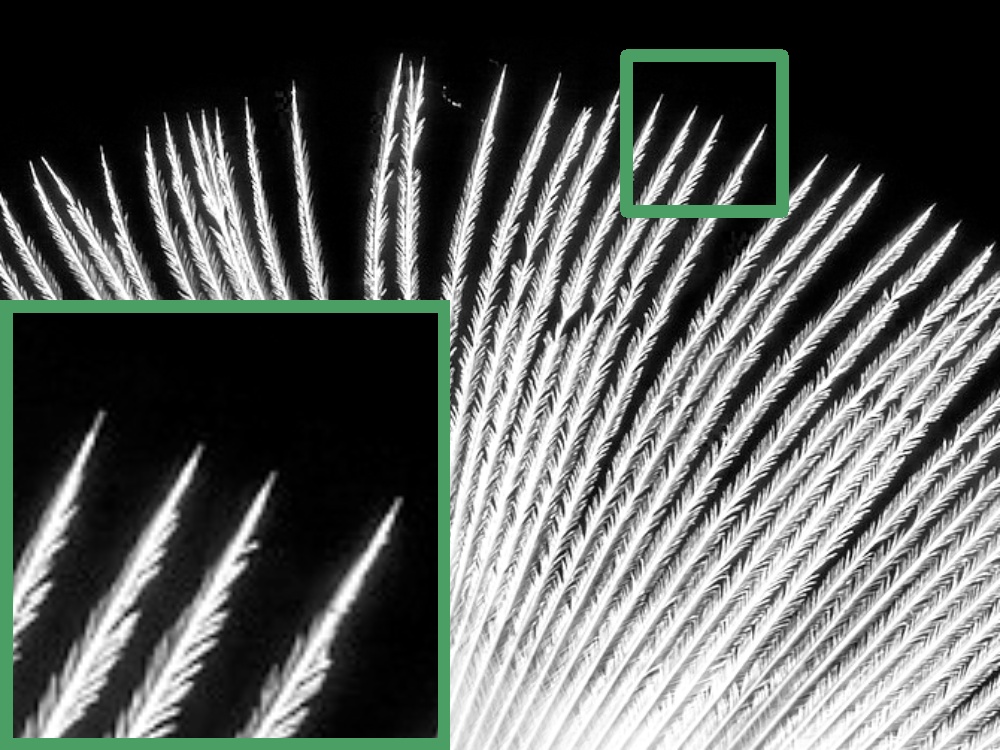}}
            \fbox{\includegraphics[width=0.24\textwidth]{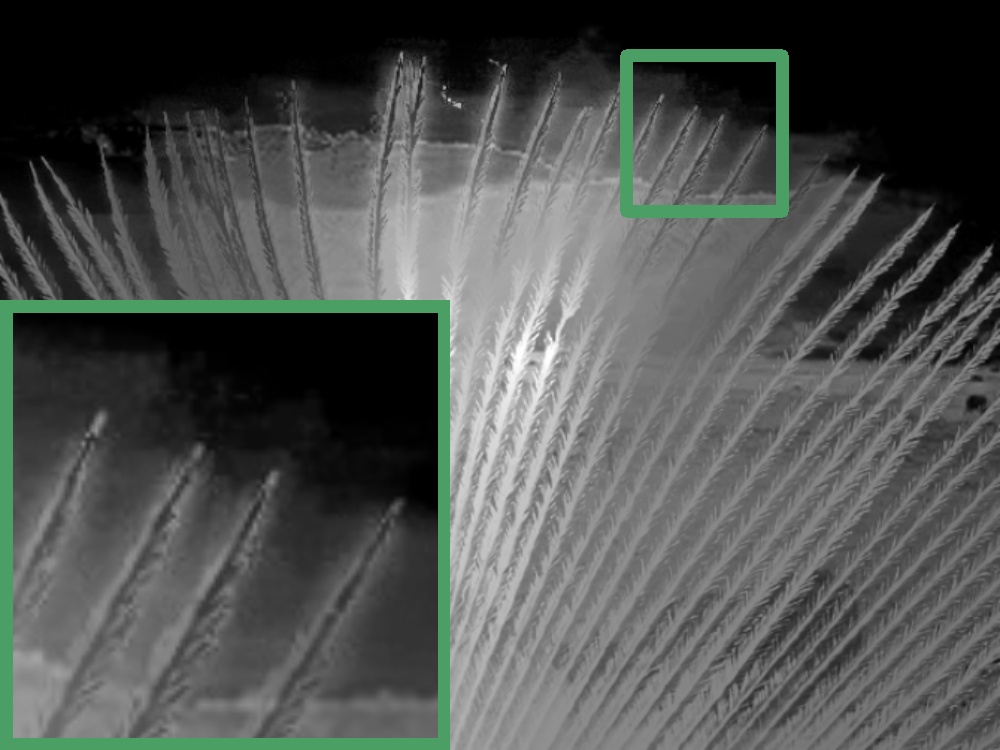}}

            \fbox{\includegraphics[width=0.24\textwidth]{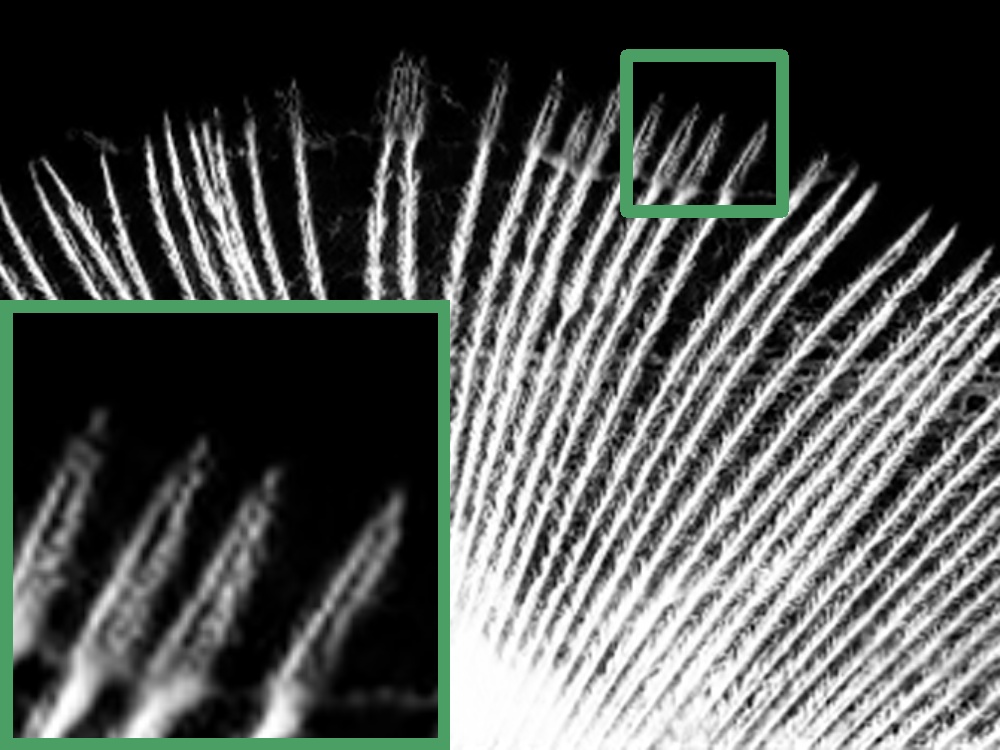}}
            \fbox{\includegraphics[width=0.24\textwidth]{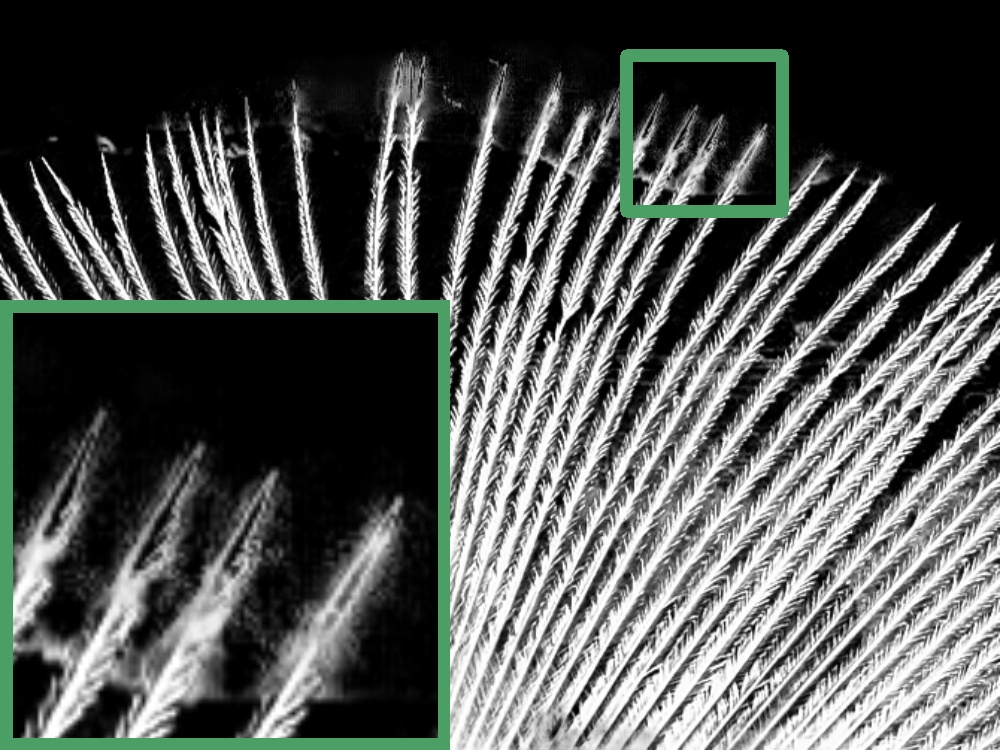}}
            \fbox{\includegraphics[width=0.24\textwidth]{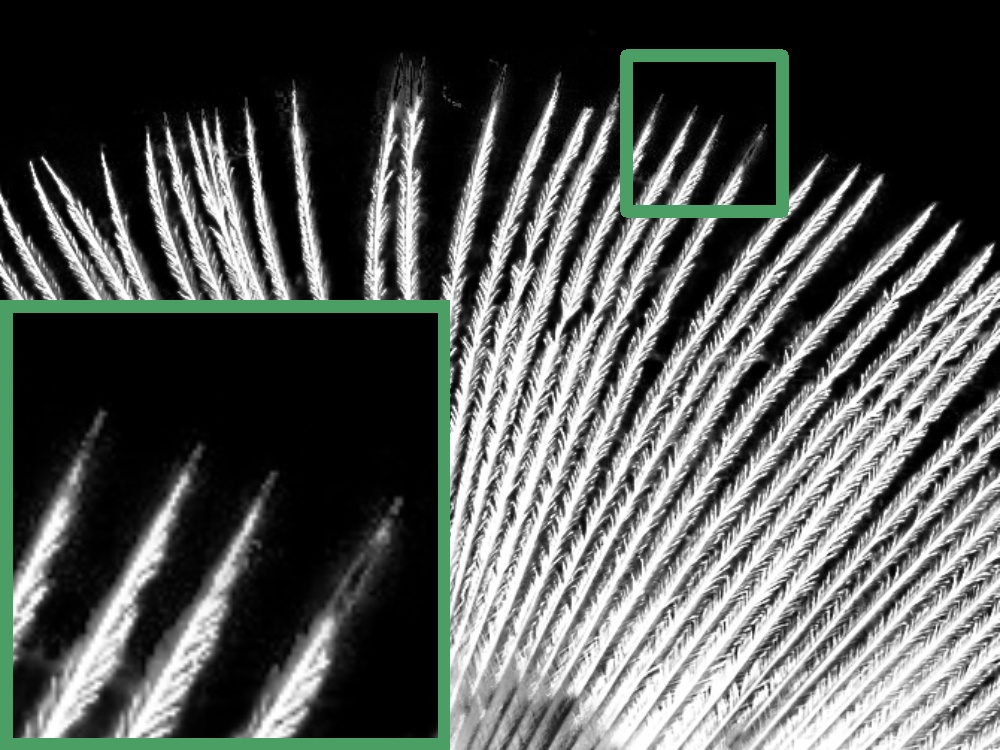}}
            \fbox{\includegraphics[width=0.24\textwidth]{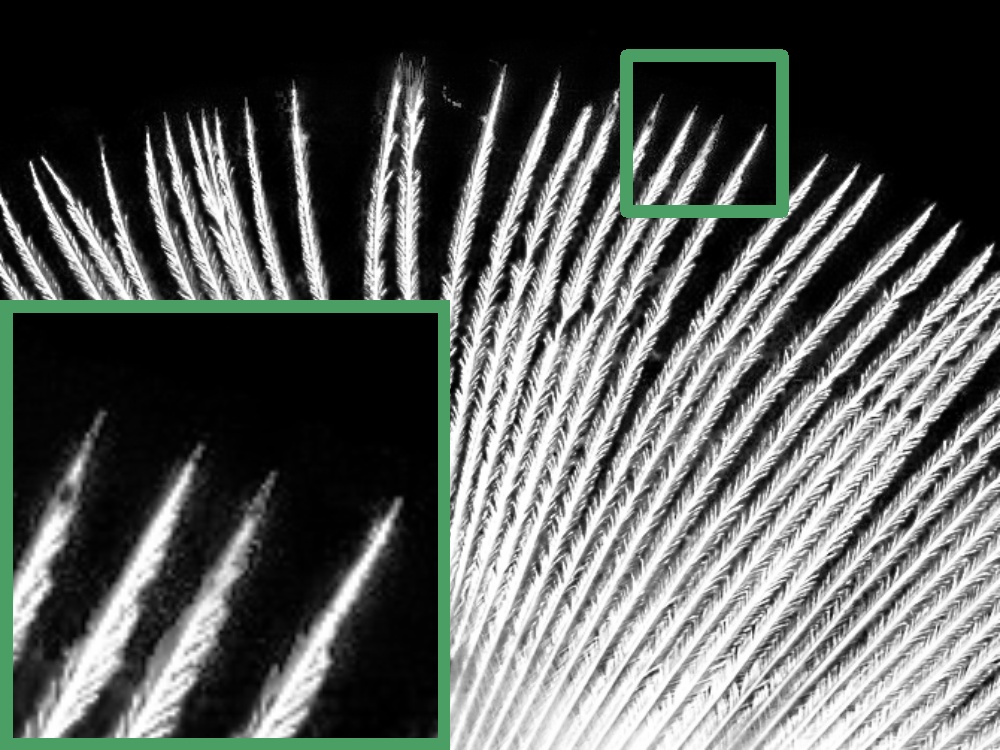}}
            \vspace{0.04in}
        \end{minipage}%
    }%

    \subfigure[``dog'']{
        \begin{minipage}[t]{0.99\textwidth}
            \centering
            \fbox{\includegraphics[width=0.24\textwidth]{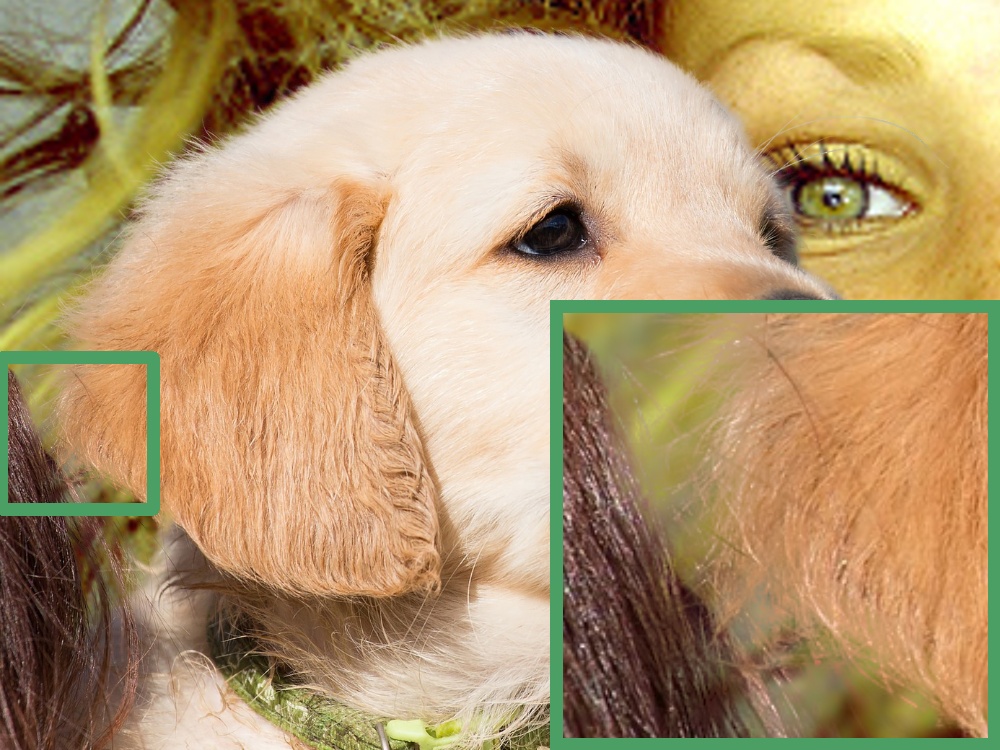}}
            \fbox{\includegraphics[width=0.24\textwidth]{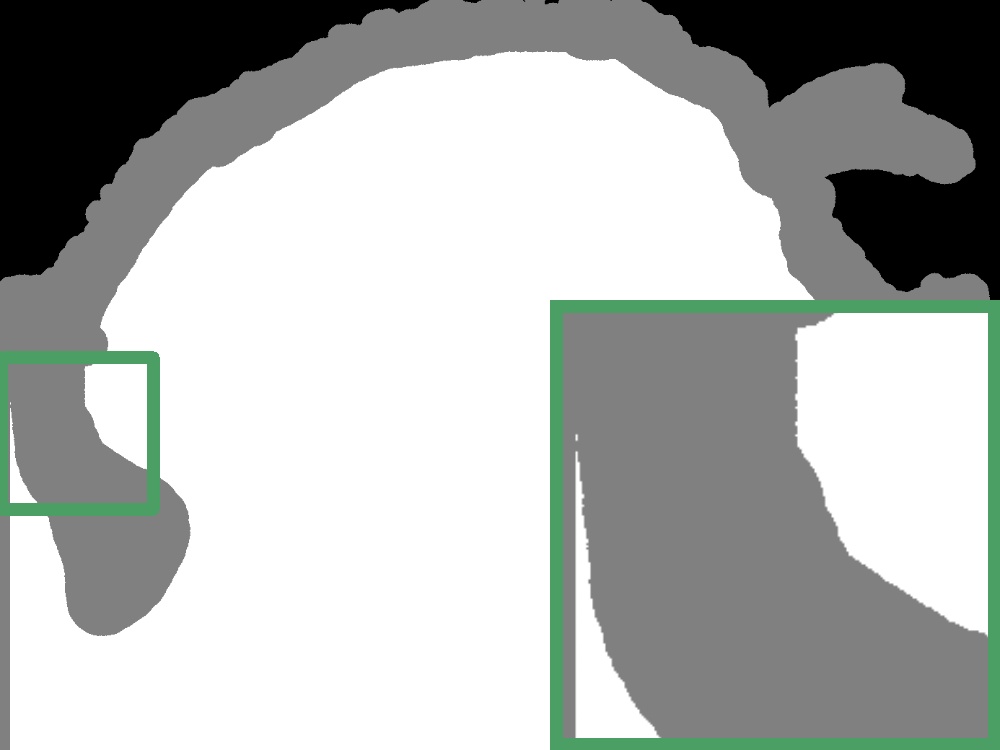}}
            \fbox{\includegraphics[width=0.24\textwidth]{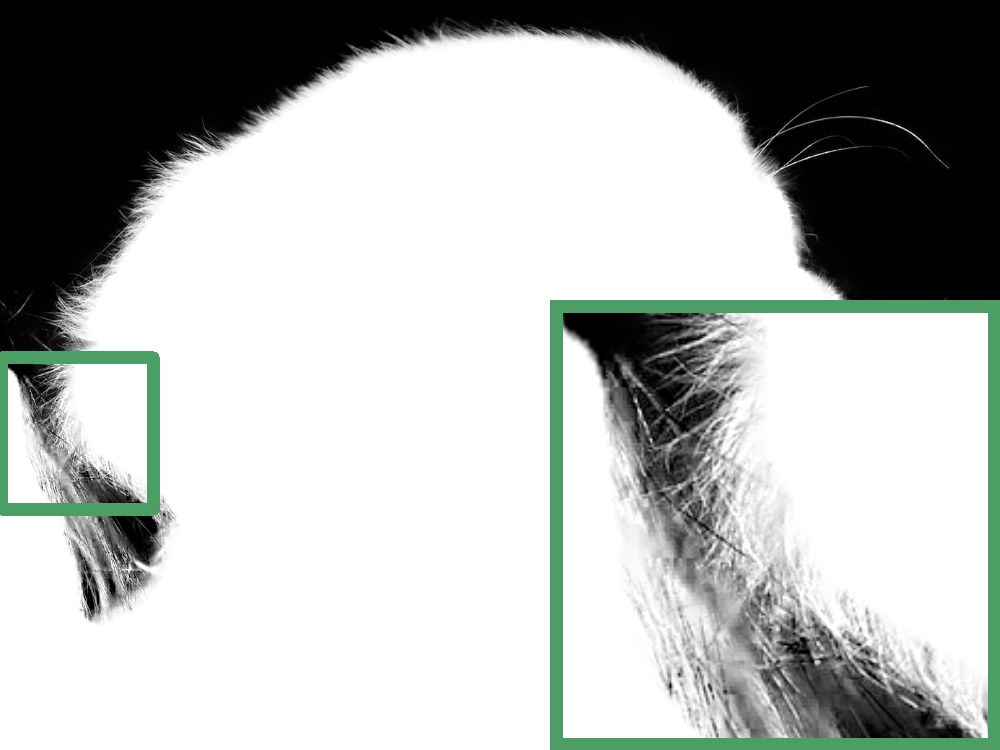}}
            \fbox{\includegraphics[width=0.24\textwidth]{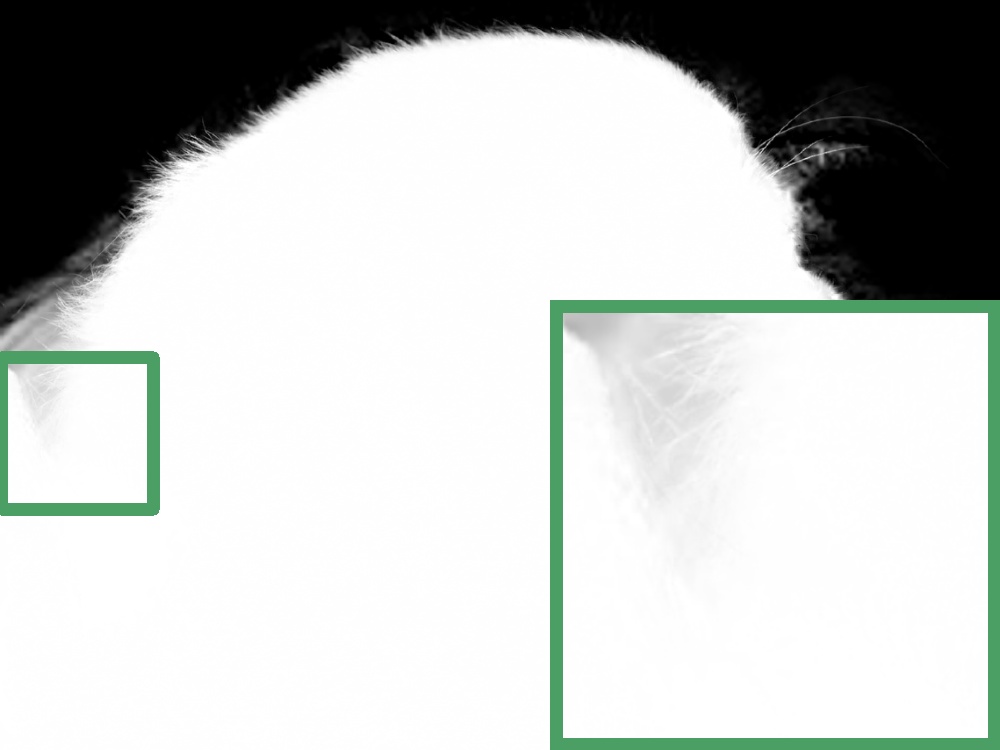}}

            \fbox{\includegraphics[width=0.24\textwidth]{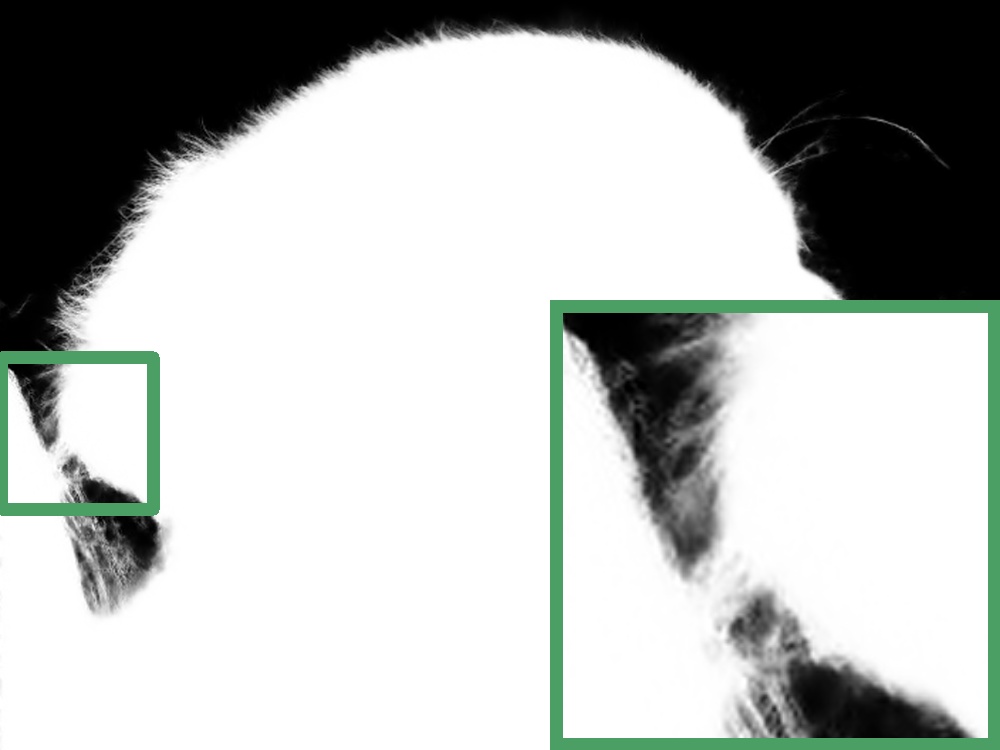}}
            \fbox{\includegraphics[width=0.24\textwidth]{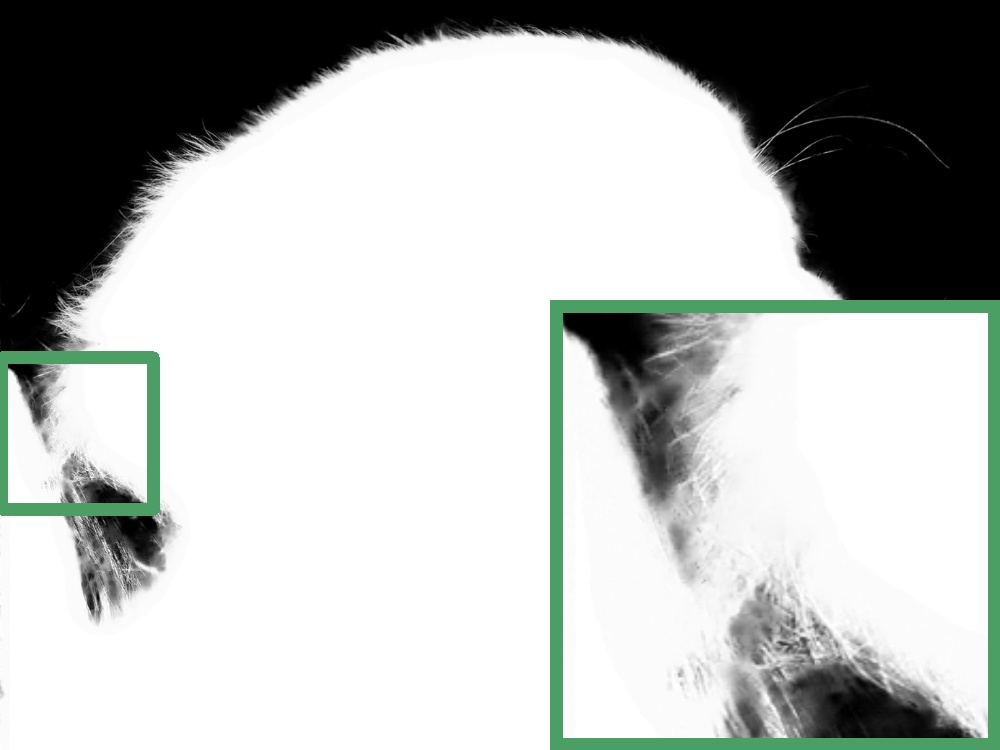}}
            \fbox{\includegraphics[width=0.24\textwidth]{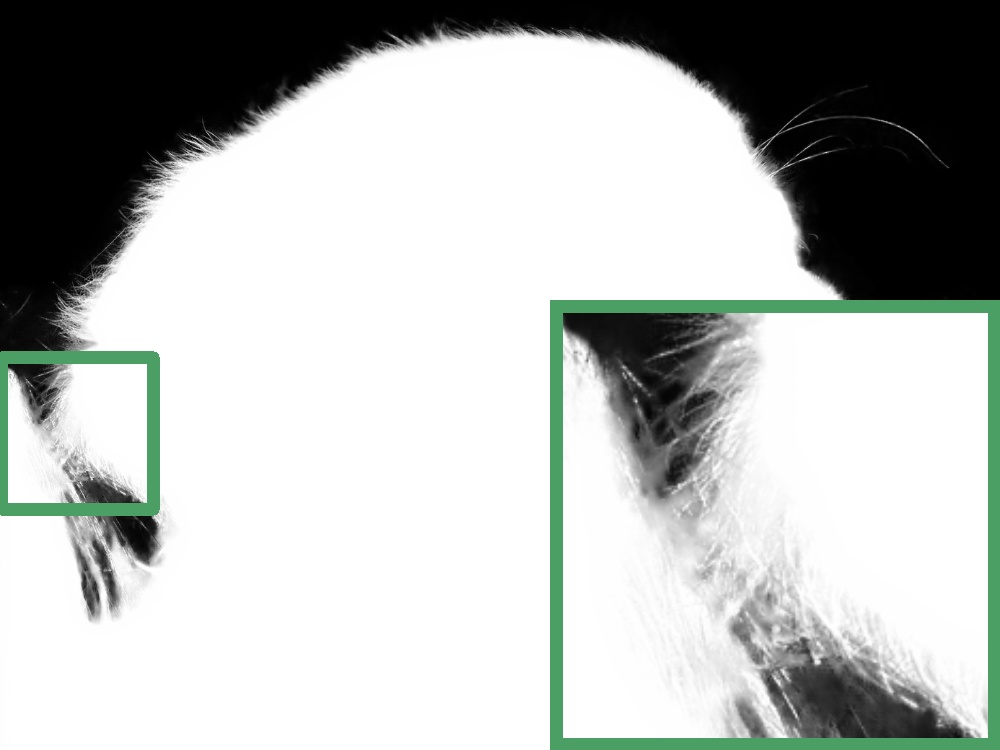}}
            \fbox{\includegraphics[width=0.24\textwidth]{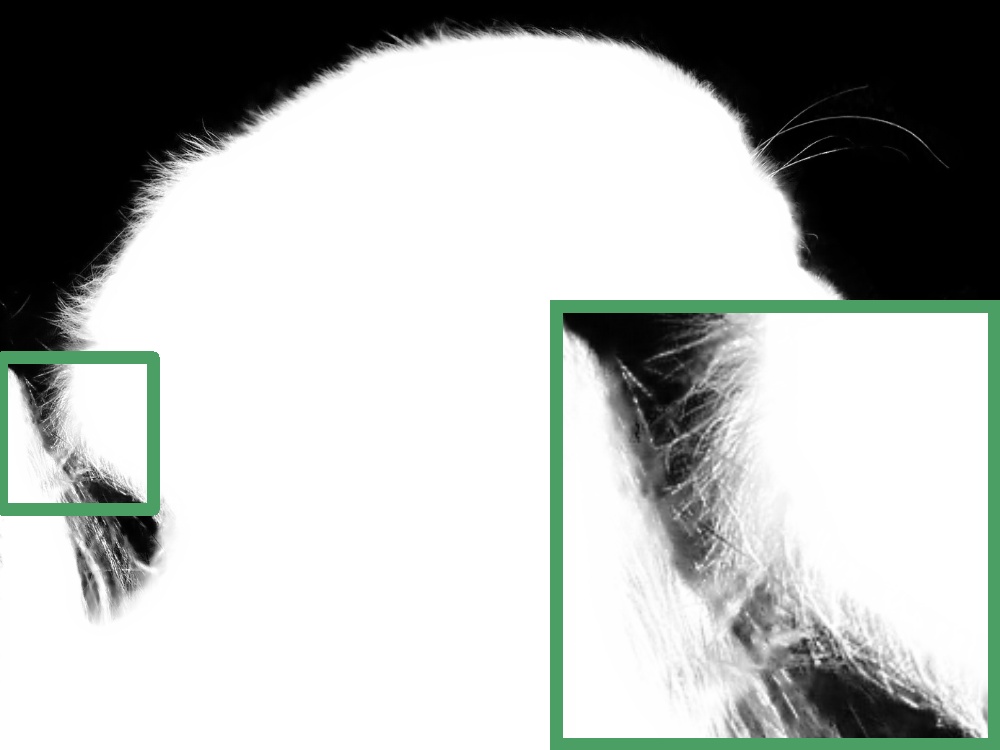}}
            \vspace{0.04in}
        \end{minipage}%
    }%
    \centering
    \caption{Example results of our method on images in Adobe Deep Image Matting Dataset. For each set of pictures, top row, from left to right: original image, trimap, ground-truth, Closed-form Matting~\cite{levin2007closed}; bottom row, from left to right: Deep Image Matting~\cite{xu2017deep}, IndexNet~\cite{lu2019indices}, our baseline, our proposed.}
    \label{fig:results}
   
\end{figure*}

The results are reported in Table \ref{tab:result}. It can also be seen that our model outperforms other state-of-the-art models in terms of SAD, MSE and Conn metrics~\cite{rhemann2009perceptually}. Example results are shown in Fig.~\ref{fig:results}. In the figure, it can be seen that our model preserved significantly more fine-grained textural details while keeping the ``absolute'' background cleaner than our baseline and other methods.

\section{Conclusions}
In this work, we propose a novel perspective of deep image matting that low-level but high-resolution features are heavily relied for recovering fine-grained details, but the downsampling operations in the very early stages of encoder-decoder architectures are harmful to these features. To prove this, we propose a deep image matting framework with two independent paths, including a dedicated downsampling-free Textural Compensate Path and an encoder-decoder based Semantic Path. The Textural Compensate Path provides more clues about fine-grained details and low-level texture features while the Semantic Path provides more high-level contextual information. Further more, we propose a novel Background Enhancement Loss and a trimap generation method to endow the model with more robustness to trimaps with various characteristics. The experimental test shows that our proposed framework significantly enhances the performance compared to the baseline, and our model outperforms other advanced start-of-the-art models in terms of SAD, MSE and Conn metrics on the Composition-1k dataset.

{\small
\bibliographystyle{ieee_fullname}
\bibliography{egbib}
}

\end{document}